\pdfoutput=1

\documentclass[11pt]{article}

\usepackage[preprint]{acl}

\usepackage{times}
\usepackage{latexsym}

\usepackage[T1]{fontenc}

\usepackage[utf8]{inputenc}

\usepackage{microtype}

\usepackage{hyperref}       
\usepackage{url}            
\usepackage{booktabs}       
\usepackage{amsfonts}       
\usepackage{nicefrac}       
\usepackage{xcolor}         
\usepackage{tabularx}
\usepackage{makecell}

\usepackage{algorithm}
\usepackage{algorithmic}

\usepackage{graphicx}
\title{On Sensitivity of Learning with Limited Labelled Data to the Effects of Randomness: Impact of Interactions and Systematic Choices}

\author{Branislav Pecher$^\spadesuit$$^\dagger$$^\ddagger$, Ivan Srba$^\dagger$, Maria Bielikova$^\dagger$$^\ddagger$ \\
$^\spadesuit$ Faculty of Information Technology, Brno University of Technology, Brno, Czechia\\
$^\dagger$ Kempelen Institute of Intelligent Technologies, Bratislava, Slovakia\\
$^\ddagger$ Slovak.AI, Bratislava, Slovakia\\
\texttt{\{branislav.pecher, ivan.srba, maria.bielikova\}}@kinit.sk}

\begin{document}

\maketitle

\begin{abstract}
  While learning with limited labelled data can effectively deal with a lack of labels, it is also sensitive to the effects of uncontrolled randomness introduced by so-called randomness factors (i.e., non-deterministic decisions such as choice or order of samples). We propose and formalise a method to systematically investigate the effects of individual randomness factors while taking the interactions (dependence) between them into consideration. To this end, our method mitigates the effects of other factors while observing how the performance varies across multiple runs. Applying our method to multiple randomness factors across in-context learning and fine-tuning approaches on 7 representative text classification tasks and meta-learning on 3 tasks, we show that: 1) disregarding interactions between randomness factors in existing works led to inconsistent findings due to incorrect attribution of the effects of randomness factors, such as disproving the consistent sensitivity of in-context learning to sample order even with random sample selection; and 2) besides mutual interactions, the effects of randomness factors, especially sample order, are also dependent on more systematic choices unexplored in existing works, such as number of classes, samples per class or choice of prompt format.
\end{abstract}

\section{Introduction}
Learning with limited labelled data, such as in-context learning, fine-tuning or meta-learning, is an umbrella term for approaches designed to work when enough labels are lacking. Although such approaches can effectively deal with limited labels, they were observed to be notably sensitive to the effects of uncontrolled randomness. Such randomness is introduced by the \textit{randomness factors}, which represent the non-deterministic decisions in the training process, such as order of samples, the model initialisation or sample choice~\cite{pham_problems_2021, gundersen2022sources_of_irreproducibility, pecher2024survey}.

The randomness in the training process can have massive impact, leading to large deviation in the performance over multiple training runs. In-context learning was found to be sensitive to the order of samples, where changing only order of samples leads from state-of-the-art predictions to random guessing~\cite{lu-etal-2022-fantastically, zhao_calibrate_2021}. Similarly, repeating fine-tuning and evaluation multiple times with different random seeds can result in smaller language models outperforming their larger counterparts~\cite{dodge_fine-tuning_2020}. 
If the randomness is not properly addressed, it can have non-negligible negative consequences even with enough labelled samples~\cite{reimers-gurevych-2017-reporting, mccoy-etal-2020-berts}. It was identified as a major obstacle to reproducibility~\cite{albertoni2023reproducibility} that can prohibit objective comparison and cause a method to be incorrectly denoted as state-of-the-art only based on more favourable chance~\cite{reimers-gurevych-2017-reporting}. The uncontrolled randomness can also unintentionally (but, unfortunately, also intentionally by cherry-picking) create an imaginary perception of research progress.

A lot of focus is dedicated to investigating and mitigating the effects of randomness and sensitivity of learning with limited labelled data~\cite{mosbach_stability_2021, pecher2024survey}, especially for in-context learning~\cite{lu-etal-2022-fantastically, zhao_calibrate_2021, chang-jia-2023-data, li-qiu-2023-finding, koksal-etal-2023-meal}. However, the existing research is often limited in its extent (in terms of randomness factors, approaches or settings) and at times leads to contradictory or inconsistent results. For example, in-context learning was believed to be consistently sensitive to the order of the randomly selected samples~\cite{lu-etal-2022-fantastically, zhao_calibrate_2021}, however, it was later observed that this sensitivity disappears when a more sophisticated sample selection strategy is used~\cite{zhang-etal-2022-active, chang-jia-2023-data, li-qiu-2023-finding}.  

We argue that the observed inconsistencies are caused by disregarding the \textit{interactions between randomness factors}, which leads to incorrectly attributing the performance deviations to different \textit{randomness factors}. Such interactions are so far partially or even completely overlooked in the existing works, resulting in the misleading findings. In addition, we hypothesise that the sensitivity of in-context learning is not only affected by the interactions with other factors but also by other \textit{systematic choices}, which are not thoroughly controlled and explored in the existing works, such as the number of classes, shots per class and prompt format.

Our main contributions are as follows\footnote{To support replicability and extension of our results, we openly publish the source code of our proposed investigation method and experiments for determining the factor importance at \url{https://github.com/kinit-sk/L3D-sensitivity-investigation}}:
\begin{itemize}
    \item We propose a novel method for investigation of \textit{randomness factors}' effects that, in contrast to the existing works, is thoroughly formalised and explicitly addresses \textit{interactions} between them by mitigating the effects of other non-investigated factors. In addition, it measures the relative importance of factors, by calculating what fraction of the overall deviation in the performance (estimated by a golden model) the investigated factor contributes in comparison to all other factors, which allows for more in-depth analysis across factors, models, datasets and experimental settings.
    \item Using the proposed method, we investigate 5 \textit{randomness factors} and their effects on in-context learning and fine-tuning across 7 representative text classification datasets, and meta-learning across 3 datasets. The results show that the in-context learning models are not consistently sensitive to the order of samples, confirming our hypothesis that the interactions play a role in the incorrect attribution of the effects of randomness factors.
    \item We further analyse how the more systematic choices influence the importance of the randomness factors. We find the following key insights: 1) predicting a higher number of classes leads to increased importance of sample order for in-context learning and reduced importance of sample order and model initialisation for fine-tuning approaches; 2) increasing the number of in-context samples reduces the importance of sample selection while having no consistent effect on the importance of sample order; and 3) the choice of prompt format has a significant impact on the importance of different factors, with larger models showing lower sensitivity to this choice.
\end{itemize} 

\section{Related Work}

The main strategy for investigating the effects of randomness factors is to repeat the training and evaluation multiple times, changing specific non-deterministic decisions of the training, such as changing what data is used and observing the change in results (i.e., \textit{Random strategy})~\cite{mccoy-etal-2020-berts, dodge_fine-tuning_2020, bouthillier2019unreproducible, agarwal_sensitivity_2021}. As such investigation may be affected by the interactions with other factors, another possibility is to perform the investigation by fixing all the other factors to a specific state (i.e., \textit{Fixed strategy}), either chosen randomly~\cite{boquet_reproducibility_2019, pham_problems_2021, zhao-etal-2021-closer} or as a result of a mitigation strategy~\cite{li-qiu-2023-finding, chang-jia-2023-data}. Another investigation strategy is to vary all the investigated factors at the same time and then decouple their effects in evaluation~\cite{dodge_fine-tuning_2020, bouthillier2021accounting, sellam_multiberts_2022, weber-etal-2023-mind, webson-pavlick-2022-prompt}, which accounts for the interactions but introduces a significant increase in computation costs~\cite{bouthillier2021accounting}.

Majority of the focus on investigating and mitigating effects of randomness is on in-context learning, which was found to be especially sensitive to the choice of samples~\cite{liu-etal-2022-makes, zhang-etal-2022-active, chang-jia-2023-data, li-qiu-2023-finding, koksal-etal-2023-meal} and their order~\cite{lu-etal-2022-fantastically, zhao_calibrate_2021, nguyen2023context}. However, it was observed that the sensitivity to sample order disappears when using a more sophisticated sample selection strategy instead of random selection~\cite{zhang-etal-2022-active, chang-jia-2023-data, li-qiu-2023-finding}, hinting at interactions between these factors that may lead to inconsistent results. In addition, the performance of in-context learning was found to be sensitive to more systematic choices as well~\cite{weber-etal-2023-mind}, such as the format of the prompt~\cite{sclar2023quantifying, voronov2024mind} or number of shots~\cite{liu-etal-2022-makes, mavromatis2023examples}. However, the impact of these systematic choices on the effects of randomness factors is not thoroughly investigated. Besides order of in-context examples, large language models were found to be especially sensitive to the order of choices in multi-choice question answering~\citep{zong2024foolvisionandlanguage, wei2024unveiling}. Although the remaining approaches and randomness factors receive only limited focus, they were still found to be sensitive to the effects of randomness, such as fine-tuning being sensitive to the random seeds (that influence model initialisation and order of samples)~\cite{dodge_fine-tuning_2020, mccoy-etal-2020-berts, mosbach_stability_2021, zhao-etal-2021-closer, zhong-etal-2021-larger}, meta-learning being sensitive to the choice of adaptation samples or how they are split into tasks~\cite{agarwal_sensitivity_2021, setlur_is_2021, cioba_how_2022, ye-etal-2021-crossfit}, or the overall machine learning being sensitive to factors such as the impact of framework and hardware implementation~\cite{boquet_reproducibility_2019, pham_problems_2021}, or the data split~\cite{bouthillier2019unreproducible, bouthillier2021accounting}.

In majority of the cases, the effects of randomness are evaluated based on a single aggregated metric from multiple runs (e.g., mean, standard deviation, or the difference between best and worst run), with the importance being determined in a binary fashion by comparing this metric to a threshold, which allows only for simple analysis~\cite{mccoy-etal-2020-berts, ye-etal-2021-crossfit, zhang-etal-2022-active}). A slightly more nuanced analysis is possible only in specific cases, where statistical approaches are used, such as grouping runs and aggregating on group level~\cite{dodge_fine-tuning_2020}, decoupling interactions~\cite{boquet_reproducibility_2019} or estimating distribution from lower number of training runs~\cite{sellam_multiberts_2022}. However, almost no studies analyse the importance of the effects in a way that would allow for easy comparison across different settings, such as what fraction of the overall variance the specific factor contributes.

We build on the ideas from the existing works, mainly from~\citep{dodge_fine-tuning_2020, bouthillier2021accounting, zhao-etal-2021-closer}, to explicitly take interactions into consideration and analyse the importance of the found effects. In addition, we fill the identified research gap by analysing the impact of more systematic choices on the randomness factors.

\section{Investigation of Randomness while Taking Interactions into Consideration}
\label{sec:methodology}

\begin{algorithm}[!tbh]
\caption{Investigate randomness factor with interactions and determine its importance} \label{alg:investigation}
\begin{algorithmic}[1]
\REQUIRE $K$: number of randomness factors
\REQUIRE $\mathbb{RF}$: set of randomness factors to consider
\REQUIRE $\mathbb{C}_{1}, \mathbb{C}_{2}, ..., \mathbb{C}_{K}$: set of configurations for each factor

\STATE Select randomness factor $i$ to investigate from $\mathbb{RF}$
\STATE Set $\mathbb{IFC}_i = \mathbb{C}_i$
\STATE Set $\mathbb{MFC}_i = \mathbb{C}_1 \times ... \times \mathbb{C}_{i-1} \times \mathbb{C}_{i+1} \times ... \times \mathbb{C}_K$

\FORALL{$m$ in $\mathbb{MFC}_i$}
    \FORALL{$n$ in $\mathbb{IFC}_i$}
        \STATE Determine model performance $r_{m, n}$ by training and evaluating the model using $m$ and $n$
    \ENDFOR
    \STATE Calculate $p\_mean_{m} = \overline{r_{m, *}}$
    \STATE Calculate $p\_std_{m} = std(r_{m, *})$
\ENDFOR
\STATE Calculate contributed standard deviation $c\_std = \overline{p\_std_{*}}$
\STATE Calculate mitigated standard deviation $m\_std = std(p\_mean_{*})$

\vspace*{0.50\baselineskip}

\STATE Set $\mathbb{GMC}_i = \mathbb{C}_1 \times \mathbb{C}_2 \times ... \times \mathbb{C}_{K-1} \times \mathbb{C}_K$

\FORALL{$g$ in $\mathbb{GMC}$}
    \STATE Determine golden model performance $r_g$ by training and evaluating the model using $g$
    \ENDFOR

\STATE Calculate overall golden model standard deviation $gm\_std = std(r_*)$

\vspace*{0.50\baselineskip}

\STATE Calculate importance score of the investigated factor $importance = (c\_std - m\_std) / gm\_std$
\IF{$importance > 0$}
    \STATE Effects of factor $i$ considered important
\ENDIF
\end{algorithmic}
\end{algorithm}

We propose a new method for investigating the effects of any \textit{randomness factor} that takes the interactions between the effects of other factors into consideration, and which is designed to measure the importance of the found effects. The steps of the method are compiled in Algorithm~\ref{alg:investigation}, with further supplementary details included in Appendix~\ref{appendix:investigation-illustration}).

\paragraph{Setup.} First, a set $\mathbb{RF}$ ($|\mathbb{RF}| = K$) is defined, which includes all the factors that will be considered in the investigation. Each \textit{randomness factor} is characterised by a set of its \textit{randomness factor configurations}, $\mathbb{C}_j$, specifying all the possible states the factor can appear in. For example, the different permutations of samples represent the \textit{configurations} of the data order randomness factor. For each factor $i$, the \textit{investigated factor configurations} set $\mathbb{IFC}_i$, containing the configurations used for the investigation, is defined as $\mathbb{IFC}_i = \mathbb{C}_i$, and the \textit{mitigated factor configurations} set $\mathbb{MFC}_i$, containing the joint configurations of the remaining randomness factors, is defined as a cross product between all the sets of \textit{randomness factor configurations}, except for the investigated randomness factor ($\mathbb{C}_i$):
\begin{equation}
    \mathbb{MFC}_i = \mathbb{C}_1 \times ... \times \mathbb{C}_{i-1} \times \mathbb{C}_{i+1} \times ... \times \mathbb{C}_K
\end{equation}

\paragraph{Investigating effects.} At its core, the investigation of factor $i$ is done by observing how the performance changes across the different configurations the randomness factor can appear in. In a single \textit{investigation run}, the training and evaluation of a model is repeated $N$ times ($N = |\mathbb{IFC}_i|$), each time with a different configurations $n$ of the factor $i$, while keeping the configurations of the remaining factors fixed to a randomly chosen configuration $m$ from $\mathbb{MFC}_i$. For each repeat, the model performance ($r_{m,n}$) is determined. The standard deviation $p\_std_{m}$ (called partial standard deviation) across these $N$ runs ($p\_std_{m} = std(r_{m, *})$) represents the effects of the investigated \textit{randomness factor} that are still affected by the interactions.

\paragraph{Mitigating interactions.} To remove the effects of other randomness factors, the \textit{investigation run} is repeated multiple ($M$) times each time with a different fixed configuration $m$ from $\mathbb{MFC}_i$. Each such repeat is called \textit{mitigation run} and results in a separate partial standard deviation. After performing enough \textit{mitigation runs} (i.e., searching through enough configurations $m$ of the non-investigated randomness factors), the partial standard deviations ($p\_std_{m}$) are averaged to produce the \textit{contributed standard deviation} ($c\_std = \overline{p\_std_{*}}$), which represents the final adjusted effects of the investigated factor $i$ (i.e., it represents the deviation the investigated randomness factor contributes to the overall deviation in results).

\paragraph{Calculating importance score.} To assess the importance of the factor, the \textit{contributed standard deviation} is compared with two additional values: 1) \textit{mitigated standard deviation} ($m\_std$); and 2) \textit{golden model standard deviation} ($gm\_std$). The \textit{mitigated standard deviation} represents the joint effects of all the non-investigated \textit{randomness factors} (i.e., standard deviation contributed by non-investigated factors). To obtain this value, a \textit{partial mean} ($p\_mean_m$) is calculated for each investigation run, which represents the expected average model performance for the given combination of configurations of the non-investigated factors. The mitigated standard deviation is then calculated as the standard deviation across these partial means ($m\_std = std(p\_mean_{*})$).

 The \textit{golden model standard deviation} ($gm\_std$) represents an objective estimate of the deviation in the model performance. To get this estimate, a \textit{golden model configuration} set $\mathbb{GMC}$ ($|\mathbb{GMC}| = L$) is defined, as a cross product between the sets of all the randomness factor configurations:
\begin{equation}
    \mathbb{GMC} = \mathbb{C}_1 \times \mathbb{C}_2 \times ... \times \mathbb{C}_{K-1} \times \mathbb{C}_K
\end{equation}
Afterwards, a model is trained and evaluated $L$ times each time with different configuration $g$ from $\mathbb{GMC}$, the model performance $ r_g $ is determined and the standard deviation across these runs represents the \textit{golden model standard deviation} $gm\_std$.

The final importance score of the factor is defined as the portion of the golden model standard deviation the investigated factors contribute over the non-investigated ones ($importance = (c\_std - m\_std) / gm\_std$). Any randomness factor with an importance value over $0$ is considered important, as it contributes the same amount of deviation as all the remaining factors combined. The size of the score determines the relative importance between the factors (e.g., factor with importance score of 0.6 is more important than one with score of 0.1) and can be used for further analysis and comparison across different factors, models, datasets and experimental settings (e.g., how the importance of specific factor changes if the number of samples is increased or a different dataset is used).

\paragraph{Choosing values for parameters $N$, $M$ and $L$.} The number of \textit{investigation runs} ($N$) and the \textit{mitigation runs} ($M$) provide a trade-off between the feasibility (or computation costs) of the investigation and the precision of the results (how well the effects are estimated and interactions mitigated). Below, we provide a set of heuristics to achieve a good trade-off (and provide full method for selecting the values of the parameters in Appendix~\ref{appendix:selecting-n-m}):
\begin{enumerate}
    \item $ N, M \gg 1 $; $ N $ and $ M $ should cover a large enough number of factor configurations to sufficiently estimate the effects.
    \item $ M \ge N $; as the higher number of mitigation runs ($ M $) leads to better mitigation of the interactions, increasing value of $ M $ should be preferred over increasing the number of investigation runs ($ N $).
    \item $ L = N*M $; to guarantee the importance score is calculated from distributions of the same sizes and characteristics, the number of runs in the golden model should be equal to the overall number of runs in the investigation.
\end{enumerate}

\paragraph{Validation of the proposed method.} We evaluate the validity of the proposed method indirectly (as there is no ground-truth to compare against) using the following experiments: 1) comparing the method to two existing baselines (i.e., \textit{Random} and \textit{Fixed} investigation strategy) and evaluating the properties and benefits of our method, specifically the handling of interactions that may lead to underestimation or overestimation of the effects in specific cases, and the importance score that allows for more in-depth analysis and comparison across different experimental settings; 2) exploring the dependence of how well the effects are estimated and their interactions mitigated by our method to the number of investigation and mitigation runs, where we found that the results of our methods are stable already with a low number of runs (20 mitigation and 10 investigation runs); and 3) observing the consistency of the results and findings when applying the method to different settings (factors, approaches, datasets). The full description of the validation results is in Appendix~\ref{appendix:investigation-strategy-comparison}.

\section{Experiments}

\paragraph{Datasets.} The experiments are conducted on 7 text classification datasets composed of different tasks with different number of classes. We focus on 3 binary classification datasets from the GLUE benchmark~\cite{wang-etal-2018-glue}: \textbf{SST2}~\cite{socher-etal-2013-recursive} for sentiment classification, \textbf{CoLA}~\cite{warstadt-etal-2019-neural} for determining the grammatical acceptability of a sentence, and \textbf{MRPC}~\cite{dolan-brockett-2005-automatically} for determining the semantic equivalence relationship between two sentences. In addition, we use 4 multi-class text datasets: \textbf{AG News}~\cite{zhang2015agnews} for news classification, \textbf{TREC}~\cite{voorhees2000trec} for question classification, \textbf{DB-Pedia}~\cite{lehmann2015dbpedia} for topic classification and \textbf{SNIPS}~\cite{coucke2018snips} for intent classification.

\paragraph{Approaches.} The main focus of the investigation is on the \textbf{in-context learning} using the Flan-T5~\cite{chung2022scaling} base, LLaMA-2~\cite{touvron2023llama} 13B instruction optimised model, Mistral-7B~\cite{jiang2023mistral} and Zephyr-7B~\cite{tunstall2023zephyr}. In addition, we also focus on \textbf{fine-tuning}, using the BERT~\cite{devlin-etal-2019-bert} and RoBERTa~\cite{liu2019roberta} base models. Finally, we also investigate the \textbf{meta-learning} approaches MAML~\cite{finn2017maml}, Reptile~\cite{nichol2018reptile} and the Prototypical Networks~\cite{snell2017prototypical}, but only on the binary datasets.

\paragraph{Randomness Factors.} In the experiments, we evaluate following randomness factors: 1) \textbf{Label Selection} used to determine the samples considered as labelled during training; 2) \textbf{Data Split} used to split the data into training, validation and test sets; 3) \textbf{Data Order} that determines the order of samples in training (order of in-context examples in prompts for in-context learning, order in which samples appear in batches for fine-tuning or tasks in meta-learning); 4) \textbf{Sample Choice} (not relevant for fine-tuning) that determines the randomly chosen samples used as in-context examples for in-context learning (or adaptation samples for meta-learning); and 5) \textbf{Model Initialisation} (not relevant for in-context learning) related to the randomly initialised weights and other parameters in the models.

\paragraph{Method Setup.} For each \textit{randomness factor}, the number of the investigation runs ($N$) is set to 10, the number of mitigation runs ($M$) is set to 100 for fine-tuning, meta-learning and 20 for in-context learning. The golden model uses the same overall number of runs ($L$) (1 000 for fine-tuning and meta-learning, 200 for in-context learning). These values, selected based on an Ablation Study (included in Appendix~\ref{appendix:ablation-study}), provide a balance between the coverage of the configurations' state space and the computation costs.

\paragraph{Experimental Setup.} We focus on a setting with limited labelled data, which represents a practical real-world scenario where a limited budget requires us to choose what data we label (a common case for many NLP supervised tasks). To simulate the unavailability of labels, we randomly select 1000 train samples from a sufficiently large labelled dataset and consider only these to be labelled. Before choosing this subset of samples, each dataset is split into train and test using 80-20 split. In addition, 20\% of the labelled train samples are used as a validation set. As such, we use different training, validation and test samples across different runs. We report the performance using the F1 macro metric. If not specified otherwise, we run in-context learning in a 2-shot setting with the first prompt format from Table~\ref{tab:prompt-format}. All prompt formats and further experimental details are included in Appendix~\ref{appendix:implementation-experimental-setting}.

\begin{table}[tbh]
\begin{center}
\footnotesize
\begin{sc}
\tabcolsep=0.11cm
\begin{tabular}{@{}lrrr@{}}

\toprule
Flan-T5        & Random                  & Fixed                   & Interactions        \\ \midrule
Golden Model   & 2.244                   & 2.244                   & 2.244       \\
Label Select.  & (*) 2.517               & (*) 2.594               & (*) 2.128   \\
Data Split     & (*) 2.362               & (*) 2.480               & (*) 2.167   \\
Data Order     & \textbf{(*) 2.131}      & \textbf{(*) 3.014}      &     0.869   \\ 
Sample Choice  & (*) 2.370               & (*) 3.191               & (*) 2.123   \\
\midrule \midrule
Zephyr-7B      & Random                  & Fixed                   & Interactions        \\ \midrule
Golden Model   & 1.043                   & 1.043                   & 1.043       \\   
Label Select.  & (*) 1.122               & (*) 1.004               & (*) 0.863   \\
Data Split     & (*) 1.185               & \textit{0.402}          & (*) 0.664   \\
Data Order     & \textbf{(*) 1.138}      & \textbf{(*) 0.957}      &     0.456   \\ 
Sample Choice  & (*) 1.052               & \textit{0.406}          & (*) 0.744   \\
\bottomrule
\end{tabular}
\end{sc}
\end{center}
\caption{Comparison of different investigation strategies for the Flan-T5 and Zephyr-7B models on the SST2 dataset based on the F1 macro standard deviation. Factors considered important for different strategies are denoted using the (*) symbol. We observe that interactions between factors may cause some factors to have their importance overestimated (denoted in \textbf{bold}) or underestimated (denoted in \textit{italics}).}
\label{tab:comparison-strategies-icl-only}
\end{table}

\begin{figure*}[tbh]
    \centering
    \includegraphics[width=0.975\linewidth]{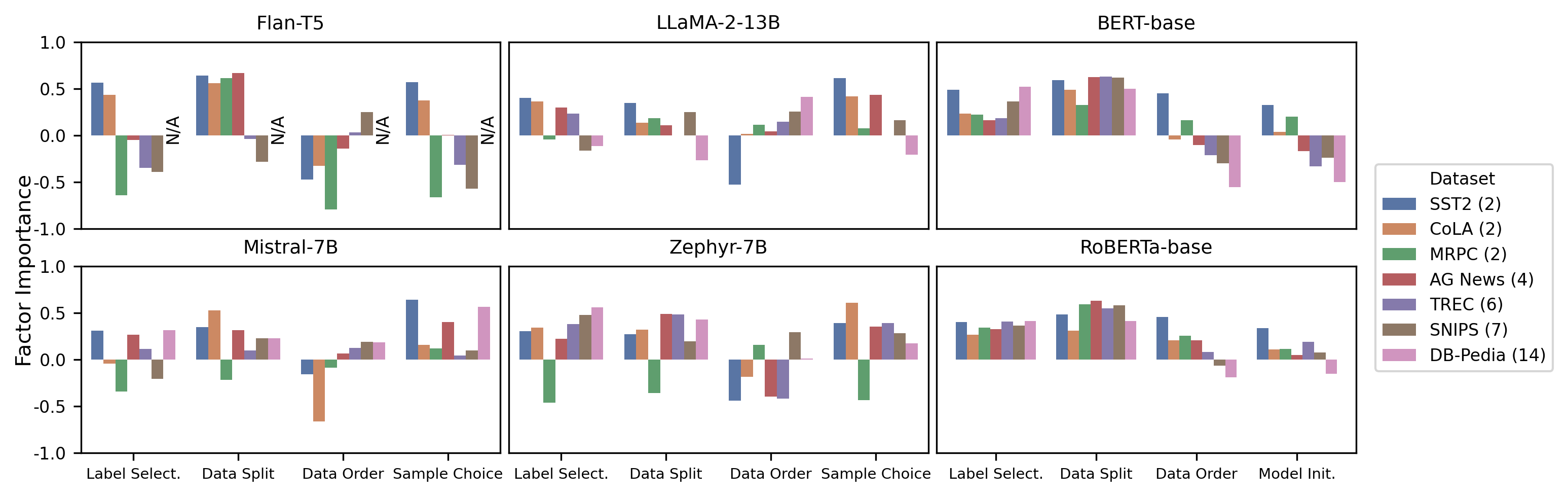}
    \caption{Importance of the investigated randomness factors for all investigated approaches and datasets while taking the interactions between factors into consideration. The legend indicates the number of classes for each dataset. As the Flan-T5 model predicts the same class for every sample on the DB-Pedia dataset, we do not include these results. Increasing the number of classes in datasets results in increased importance of the Data Order factor for in-context learning and reduced importance of Data Order and Model Initialisation for fine-tuning approaches.}
    \label{fig:rf-importance-full}
\end{figure*}

\subsection{Interactions Between Randomness Factors}

In this section, our goal is to answer the following research question: \textit{\textbf{RQ1:} How do the interactions between randomness factors affect their individual importance?} To answer this question, we compare our proposed method (\textbf{Interactions}) with the commonly used investigation strategies: 1) \textbf{Random}, which varies the overall random seed in the investigation without any constraint on the configurations of other factors; and 2) \textbf{Fixed}, where the non-investigated randomness factors are fixed to a single configuration for all runs of the investigation. For these strategies, we consider the effects of factor to be important when it contributes at least 50\% of the golden model standard deviation. The results from this comparison are shown in Table~\ref{tab:comparison-strategies-icl-only} and used for validation of our method in Appendix~\ref{appendix:investigation-strategy-comparison}.

\textbf{Effects of randomness factors may be overestimated or underestimated when interactions are not taken into consideration.} The \textit{Random} strategy leads to a deviation similar to the one from the golden model across all investigated randomness factors. Such result indicates that all randomness factors are equally important, leading to a significant importance overestimation in some cases (e.g., Data Order factor for both Flan-T5 and Zephyr models). Even though the \textit{Fixed} strategy produces more reliable results, it is still affected by the random choice of the single factor configuration (e.g., Data Order contributing deviation of $3.014$, which is much higher than the deviation of $2.244$ from the golden model). As such, we observe underestimation of the results (e.g., Sample Choice and Data Split with the Zephyr-7B model not being considered important with a deviation of $0.406$ and $0.402$) as well as overestimation (e.g., Data Order being considered important with a deviation of $3.014$ for Flan-T5 and $0.957$ for Zephyr-7B). Taking the interactions into consideration, we observe that the \textbf{Data Order randomness factor is not consistently important for in-context learning even when choosing samples randomly}, which confirms the impact of interactions on incorrect attribution of effects of different randomness factors.

\begin{figure*}[tbh]
    \centering
    \includegraphics[width=0.835\linewidth]{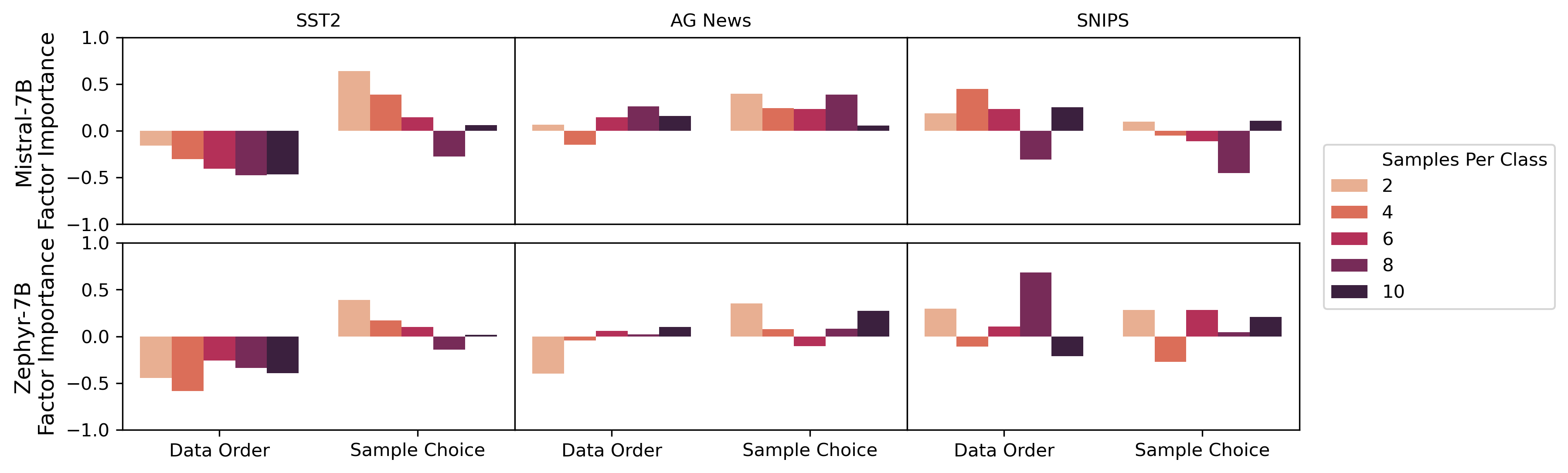}
    \caption{The change in importance of the Data Order and Sample Choice randomness factors as the number of in-context examples increases. Increasing the number of samples per class does not have a consistent effect on the importance of the Data Order factor, while the importance of the Sample Choice factor decreases.}
    \label{fig:rf-importance-shots}
\end{figure*}

\subsection{Importance of Randomness Factors}
\label{section:rf-importance}

In this section, we want to answer the following research question: \textit{\textbf{RQ2:} What randomness factors are important for different approaches for learning with limited labelled data?} We analyse the results of our method on different datasets to identify the consistently important factors. The results are included in Figure~\ref{fig:rf-importance-full} for in-context learning and fine-tuning and in Appendix~\ref{appendix:metal-results} for meta-learning.

\textbf{Sample Choice represents the most important factor for in-context learning.} For the majority of the investigated models, the Sample Choice factor is considered important for almost all of the datasets, achieving an average importance score of $0.25$ across the models and datasets. A notable exception is Flan-T5 on the multi-class datasets (average importance score of $-0.39$) or Zephyr on the MRPC dataset (importance score of $-0.43$), where the factor is not considered important.

\textbf{Importance of Data Order is dataset and model dependent for in-context learning.} Majority of the in-context learning models do not show sensitivity to the Data Order randomness factor on binary datasets (average importance score of $-0.28$). At the same time, the importance of Data Order becomes consistently higher on multi-class datasets for all models (average importance score of $0.16$), with the exception of the Zephyr-7B.

\textbf{General randomness factors, Label Selection and Data Split, show consistent importance for the majority of the models and datasets.} In case of fine-tuning, the Label Selection and Data Split randomness factors show the highest level of importance across all datasets when compared to other randomness factors (average importance score of $0.34$ and $0.52$). For in-context learning, we do not observe such consistent results, with the importance changing based on the dataset and model used. However, these factors are considered important in more than half of the cases (16 out of 27 for Label Selection and 22 out of 27 for Data Split).

\textbf{Importance of Data Order and Model Initialisation is dataset and model dependent for fine-tuning.} For the binary datasets, these factors are considered important for both models (average importance of $0.25$ for Data Order and $0.19$ for Model Initialisation). However, on the multi-class datasets, the importance of Sample Order for both models (average importance score of $-0.14$) and Model Initialisation for the BERT model (average importance score of $-0.30$ for BERT and $0.04$ for RoBERTa) drops significantly.

\subsection{Effects of Variable Number of Classes and In-Context Samples}
\label{section:classes-and-shots}

\begin{figure*}[tbh]
    \centering
    \includegraphics[width=0.835\linewidth]{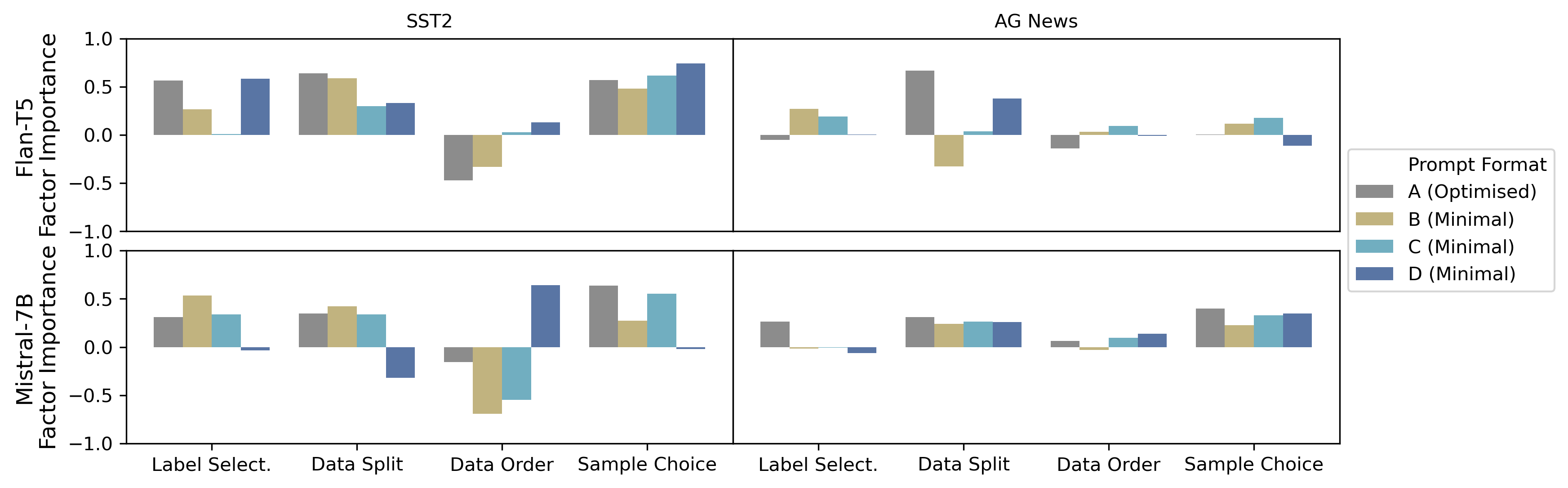}
    \caption{Effect of different prompt formats on the importance of randomness factors for in-context learning. The choice of format has a significant effect on the importance of different factors, with the minimal formats often leading to higher importance. At the same time, the larger models show lower sensitivity to prompt format.}
    \label{fig:rf-importance-format}
\end{figure*}

In this section, our focus is on answering the following research question: \textit{\textbf{RQ3:} How does the importance of data-specific randomness factors change based on the number of classes and in-context samples?} As we observe different effects of randomness factors on binary and multi-class datasets, our main focus is to determine whether the change in importance is caused by the increased number of in-context examples in the prompt, by the larger number of options that can be predicted or by a combination of both. The results from changing the number of classes are included in Figure~\ref{fig:rf-importance-full}, and from changing the number of shots for in-context learning are included in Figure~\ref{fig:rf-importance-shots}.

\textbf{The importance of Data Order randomness factor for in-context learning increases at higher number of classes.} The Data Order randomness factor is not considered important for any of the in-context learning models on the SST2 dataset, achieving importance of $-0.47$, $-0.53$, $-0.16$ and $-0.44$ respectively for the Flan-T5, LLaMA-2, Mistral and Zephyr models. On the remaining binary datasets, the importance either gradually increases (LLaMA-2 or Zephyr) or decreases (Flan-T5 and Mistral). However, on the datasets with the higher number of classes, the importance of the Data Order factor gradually increases (with the exception of the Zephyr model), achieving importance as high as $0.25$ for Flan-T5 and $0.29$ for Zephyr on the SNIPS dataset, and $0.41$ for LLaMA-2 and $0.18$ for Mistral model on DB-Pedia dataset.

\textbf{The importance of the Sample Choice for in-context learning is not consistently affected by the number of classes.} In case of Flan-T5 and Zephyr, the importance of Sample Choice gradually decreases as we increase the number of classes (from $0.57$ on SST2 to $-0.57$ on SNIPS for Flan-T5, or from $0.39$ on SST2 to $0.17$ on DB-Pedia for Zephyr). For the LLaMA-2 model, the decrease is not as consistent, with the importance being much lower on the TREC than on the SNIPS dataset. Finally, the Sample Choice randomness factor is consistently important across all datasets for the Mistral model, with no apparent tendency.

\textbf{The importance of Data Order and Model Initialisation randomness factors for fine-tuning decreases with higher number of classes.} For BERT model, we observe a gradual decrease of importance for both factors as we increase the number of classes, going from $0.45$ and $0.33$ on SST2 to $-0.55$ and $-0.50$ on DB-Pedia, respectively for Data Order and Model Initialisation. Similarly, we observe a gradual decrease of Data Order importance for RoBERTa model, going from $0.46$ on SST2 to $-0.19$ on DB-Pedia. However, Model Initialisation does not show a consistent tendency for RoBERTa, with the importance staying approximately the same across the majority of the datasets.

\textbf{Number of in-context samples has no consistent effect on the importance of Data Order.} The importance of Data Order remains consistent, or even is lowered, across all models, datasets and number of shots per class. On the other hand, \textbf{increasing the number of shots reduces the importance of Sample Choice factor} for all models and datasets. For example, the importance of Sample Choice for Zephyr drops from $0.39$ on 2-shot setting to $0.02$ on 10-shot setting on the SST2 dataset.

\subsection{Impact of Prompt Format}
\label{section:prompt-format}

In this section, we aim to answer the following research question: \textit{\textbf{RQ4:} How does the prompt format affect the importance of randomness factors?} As the previous works observed a significant sensitivity of in-context learning to prompt format our goal is to investigate whether such sensitivity affects the importance of randomness factors as well. To achieve this, we compare our optimised prompt format (Format A) with 3 minimal prompt formats (Formats B, C and D) defined in~\citep{li-qiu-2023-finding, gao-etal-2021-making, koksal-etal-2023-meal}. All the prompt formats are described in detail in Table~\ref{tab:prompt-format} in Appendix~\ref{appendix:implementation-experimental-setting}. The results from the investigation are illustrated in Figure~\ref{fig:rf-importance-format}, with full results included in Appendix~\ref{appendix:results}.

\textbf{Minimal formats lead to significant changes in the importance of randomness factors over the optimised format.} The Data Order randomness factor shows the highest sensitivity to the prompt format, becoming significantly important in many cases even when the interactions are taken into consideration. At the same time, Sample Choice is not as sensitive to the prompt format. The remaining randomness factors, Label Selection and Data Split, are affected only when using specific formats -- using the last format, we observe a significant change in the importance of these randomness factors across all models and datasets. The \textbf{larger models}, Mistral and Zephyr, \textbf{show lower sensitivity to prompt format change}, as the importance of all randomness factors remains consistent across formats. On the other hand, in case of the Flan-T5 model, the importance of randomness factors changes significantly across different formats.

\subsection{Discussion}

Besides understanding the sensitivity, an important aspect is predicting a good configurations of the most important randomness factors to guarantee stability and generalisability of the approaches. As opposed to the hyperparameter tuning, finding this configuration is not as straightforward as it is not so systematic. First, as we show in this work, the importance and the best configuration is strongly affected by the interactions between randomness factors and other systematic choices. Second, there is no metric that can serve as an estimate for the quality of the configuration besides the observed performance -- for example when finding an optimal prompt, the best performing and worst performing one can differ only in a single word~\citep{zhan2024unveiling}.

One way how to determine the optimal configuration are the mitigation strategies that recently started to attract a research attention (see the recent survey on stability of learning with limited labelled data~\citep{pecher2024survey} for more information). While the mitigation strategies are often factor-specific, such as sample selection strategies for in-context learning~\citep{li-qiu-2023-finding, chang-jia-2023-data, koksal-etal-2023-meal, pecher2024automatic}, also more general strategies based on ensembling and further model training have been developed~\citep{pecher2024fighting, pezeshkpour2023large, summers_nondeterminism_2021, wang-etal-2023-two, allingham23simple, voronov2024mind}. Uncovering the most important randomness factors through systematic investigation and design of new and effective mitigation strategies to reduce the sensitivity to the effects of randomness is an important future directions of the field~\citep{pecher2024survey, liu2023pre}.

\section{Conclusion}

In this work, we have proposed a novel method that explicitly takes interactions between different \textit{randomness factors} into consideration by mitigating the effects of the other, non-investigated, randomness factors. In addition, our method is designed to determine the importance of the investigated randomness factor by measuring what fraction of the overall deviation of the model (represented using a golden model) it contributes over the mitigated randomness factors, allowing for in-depth analysis across experimental settings.

Applying our proposed method to investigate the effects of \textit{randomness factors} on in-context learning, fine-tuning and meta-learning, we confirm our hypothesis that interactions between randomness factors may cause incorrect attribution of effects of one factor to another, leading to inconsistent result. Contrary to the previous works, after taking interactions into consideration, we do not observe a consistent sensitivity of in-context learning approaches to the sample order even when choosing samples at random. Instead, we observe that the importance of randomness factors, especially the sample order, is affected by the interactions with other factors and by the systematic choices such as the number of predicted classes, the number of samples per class and the choice of prompt format.

The proposed method can be applied to other NLP tasks as well, such as question answering, with minimal modifications. Only requirement is to define the randomness factors and their configurations, such as order of choices in the questions or the symbols used for the answers. Extending our investigation to other tasks represents an interesting potential for future work.

\section*{Acknowledgements}
This work was partially supported by the projects funded by the European Union under the EU Horizon 2020: \textit{TAILOR}, GA No. \href{https://doi.org/10.3030/952215}{952215}, by the European Union under the Horizon Europe: \textit{DisAI}, GA No. \href{https://doi.org/10.3030/101079164}{101079164} and \textit{vera.ai}, GA No. \href{https://doi.org/10.3030/101070093}{101070093}, and by the EU NextGenerationEU through the Recovery and Resilience Plan for Slovakia under the project No. 09I03-03-V03-00020.

Part of the research results was obtained using the computational resources procured in the national project \textit{National competence centre for high performance computing} (project code: 311070AKF2) funded by European Regional Development Fund, EU Structural Funds Informatization of Society, Operational Program Integrated Infrastructure.

\section*{Limitations}
The effects of randomness factors are investigated on a selection of models from different approaches for learning with limited labelled data. However, in order to provide a more extensive and in-depth analysis of the interactions and the more systematic choices without a significant increase in computation costs, the effects are investigated on models of smaller sizes -- we use the base versions of BERT, RoBERTa and Flan-T5 models, and a 4-bit quantised versions of the LLaMA-2-13B, Mistral-7B and Zephyr-7B models. As such, the observed effects may not be as representative for larger models. However, similar to related work, we observed the larger models to be more susceptible to the effects of randomness and so the results of our investigation may underestimate the importance of different factors (instead of their over-estimation).

The number of investigation and mitigation runs used in our investigation is selected based on an ablation study (in Appendix~\ref{tab:ablation-reducing}). In addition, following related work (e.g.,~\cite{gao-etal-2021-making, chang-jia-2023-data, sclar2023quantifying, li-qiu-2023-finding, koksal-etal-2023-meal}) and the results of our ablation study, we also evaluate each run using only 1 000 test samples. In both cases, the decision represents a trade-off between the reliability of the results and their feasibility. Although this represents an optimal trade-off, increasing the number of runs and the number of test samples could potentially lead to better estimation of the effects and mitigation of their interactions (especially on larger datasets), but at the cost of significant computation costs increase. At the same time, the number of investigation and mitigation runs utilised in this paper still represents a significant improvement over the existing studies, as it is a common practice to investigate the effects using very low numbers of runs. As future work, we plan to explore the possibilities for effective mitigation strategies for all the \textit{randomness factors} to mitigate their effects and further reduce computation costs.

Similarly, we investigate the effects on a smaller set of training labelled samples (using only 1 000 labelled samples). This setup may lead to larger effects of randomness and lower stability for fine-tuning and meta-learning while having negligible impact on in-context learning (which works with a smaller subset of these samples). However, as this represents a real-world scenario with a limited budget, we do not consider this to be a significant limitation, as the effects of randomness factors were previously found to be significant even with the use of large datasets~\cite{mosbach_stability_2021, dodge_fine-tuning_2020, reimers-gurevych-2017-reporting}.

Even though the effects of implementation level \textit{randomness factors} (e.g., framework implementation and hardware scheduling) were previously observed to affect the models' performance, we consider their effects only partially by mitigating them as much as possible (e.g., setting CUDA to be deterministic, using the same version of libraries and the same system for all experiments). Investigating their effects fully is out of scope for this paper due to the specifics required to thoroughly explore these effects (e.g., using a single worker, deterministic implementation, or running on a single CPU thread \cite{pham_problems_2021}).

Although we perform basic prompt engineering to obtain our optimised prompt for each dataset, the prompt format could be theoretically improved using automatic prompt-tuning methods. As we have observed the impact of prompt format on the effects of different randomness factors, the use of such format may lead to different findings, especially for the Flan-T5 model. However, we still observed the main in-context learning models (Mistral-7B and Zephyr-7B) to be sufficiently robust to this change and their results should stay the same. At the same time, our main prompt was designed based on the recommendations and prompt formats from related work~\cite{sun2023pushing, li-qiu-2023-finding, gao-etal-2021-making, koksal-etal-2023-meal}, so we do not expect significant changes when using more prompts obtained through prompt-tuning.

Finally, we are not sure whether the datasets we use in our experiments have been used to train the models we use for in-context learning, which may affect our findings and results on these models. We limit this effect by using our own optimised prompt across the majority of the experiments. However, we cannot guarantee it is enough to provide unbiased results as this limitation is part of the recently recognised LLM validation crisis~\citep{li2023task} and we would need to train the model from scratch to address it properly, which is out of scope for this paper.

\bibliography{anthology,custom}

\appendix
\appendix

\section{Ethical Considerations and Impact Statement}
The experiments in this paper work with publicly available benchmark dataset GLUE, and publicly available datasets AG News, TREC, SNIPS and DB-Pedia, citing the original authors. As we were not able to determine the license for the tasks and datasets used, we opted to use them in as limited form as possible, adhering to the terms of use (no annotation of the test set) for the GLUE benchmark dataset and applying it to other datasets as well. We do not work with any personally identifiable information or offensive content and perform no crowdsourcing for further data annotation. In addition, we are not aware of any potential ethical harms or negative societal impacts of our work, apart from the ones related to the advancement of the field of Machine Learning and Learning with Limited Labelled Data, which includes the in-context learning, transfer learning, meta-learning and language model subsets. Finally, we follow the license terms for all the models we use (such as the one required for the use of the LLaMA-2 model) -- all models and datasets allow their use as part of research. It is possible the large language models we used (Flan-T5, LLaMA-2, Mistral and Zephyr) contain biases and may generate potentially offensive or harmful content. However, the authors of the models reduce this potential bias as much as possible when training the models, while at the same time we limit the output to few tokens and do not release any output of the models, which should further reduce the potential bias and negative impact.

\paragraph{Impact Statement: CO2 Emissions Related to Experiments}
\label{compute-used}

The experiments presented in this paper used significant compute resources as they required multiple training and evaluation runs of multiple models, as well as using large language models that require a lot of computation even just for the inference. Overall, the experiments were conducted using a private infrastructure, which has a carbon efficiency of 0.432 kgCO$_2$eq/kWh (default value used as the actual efficiency of our HW instance was not measured). A cumulative of 1440 hours of computation was performed on hardware of type RTX 3090 (TDP of 350W) and a cumulative of 4000 hours of computation was performed on hardware of type A100 PCIe 40GB (TDP of 250W). The hours of computation used are only a crude approximation, as the machine used was shared among multiple projects. Total emissions are estimated to be 217.73 kgCO$_2$eq (for the first set of hardware) and 432 kgCO$_2$eq (for the second set of hardware), of which 0 percents were directly offset. These estimations were conducted using the \href{https://mlco2.github.io/impact#compute}{MachineLearning Impact calculator} presented in \cite{lacoste2019quantifying}. 
Whenever possible, we tried to reduce the compute resources used as much as possible. The most compute resources were used by the large language model -- LLaMA-2, Mistral-7B and Zephyr-7B. To reduce the computation costs and resources used, we decided to evaluate the model on lower number of runs (10 investigation and 20 mitigation, resulting in 200 runs for each randomness factors) using only 1 000 test samples for evaluation. Even in this reduced evaluation, the experiments using these models used the most GPU hours. To further reduce the compute resources, we use the 4-bit quantised versions of these models, while also opting to use smaller models for the more detailed analyses and ablation studies, either in case of in-context learning (e.g., using Flan-T5 and Mistral-7B instead of LLaMA-2 for studying the impact of number of shots that significantly increased the required computation resources and inference time), but also in case of transfer learning and meta-learning (e.g., using base versions of the BERT and RoBERTa models).

\section{Additional Resources Describing the Proposed Investigation Method}
\label{appendix:investigation-illustration}

In this Appendix, we provide additional supplementary resources that should allow for easier understanding of how the method operates. The proposed investigation method is designed for investigating effects of any \textit{randomness factor}, while explicitly taking interactions with effects of other factors into consideration, and measuring the importance of the found effects.
Overall, the effects are investigated by observing how the performance changes across the different states the investigated randomness factors can appear in. To deal with the interactions, the effects of remaining randomness factors are mitigated (i.e., the deviation they contribute is reduced as much as possible). To determine the importance, we compare the contributed deviation of the investigated randomness factor with effects of other factors, and with the overall deviation from a golden model. The golden model represents the objective estimate of the performance deviation and is obtained by training a model while mitigating all the randomness factors at the same time. The final importance score is then determined as the fraction of the overall deviation (represented using the golden model) the investigated factor contributes over all the remaining, non-investigated factors. 

The following section provides a high-level overview of the method with references to the Algorithm~\ref{alg:investigation} (Appendix~\ref{appendix:investigation-highlevel}), the illustration of how the method operates and the results it computes (Appendix~\ref{appendix:investigation-full-illustration}). We also provide a method for selecting the number of investigation and mitigation runs in Appendix~\ref{appendix:selecting-n-m} (this method was used to select the samples in this paper using the Ablation Study in Appendix~\ref{tab:ablation-reducing} and to produce the heuristics at the end of Section~\ref{sec:methodology}).

\subsection{Algorithmic Description of the Method}
\label{appendix:investigation-highlevel}

To allow for better understanding of our proposed investigation method, we provide more informal description of the steps composed in Algorithm~\ref{alg:investigation}, along with references to individual lines in it and possible avenues for extension of our method. Informally our proposed method works in a following way:
\begin{enumerate}
    \item A set of randomness factors for investigation is first identified along with their configurations. In case of mitigated randomness factors, a complete set of factors and their configurations is not required to prevent introduction of biases into the results, as the randomness factors can be controlled on the group level. All the algorithmic factors (order, initialisation, model randomness, etc.) can be controlled by globally setting the seed, while the implementation/hardware level factors can be controlled using the same setup across all experiments (same library versions, architectures, GPUs, etc.). 
    \item A single investigation run is performed for a selected investigated randomness factor (repeating and evaluating training multiple times, each time with different configuration of the selected randomness factor, e.g., with different split of data, choice of data, or their order), while keeping the configuration of all other (non-investigated) randomness factors fixed. (inner loop; lines 5-7 in the Algorithm~\ref{alg:investigation}) 
    \item The method can be easily extended to investigate effect of multiple factors at the same time, by simply changing the definition of the \textit{investigated factor configuration} set, to include a cross product between the configurations of multiple factors (similarly to the \textit{mitigated factor configuration} set). (lines 2 and 3 in the Algorithm~\ref{alg:investigation})
    \item The single investigation run is evaluated to obtain partial standard deviation and partial mean. (lines 8-9 in the Algorithm~\ref{alg:investigation})
    \item The configuration of all other randomness factors is fixed to a new value and the investigation run is repeated again to mitigate the effects of non-investigated randomness factors (each such repeat is called \textit{mitigation run}). (outer loop; lines 4-10 in the Algorithm~\ref{alg:investigation})
    \item Instead of repeating multiple mitigation runs, the method can be extended to use a specific mitigation strategy (such as sample selection method for in-context learning). Using such method, the set of configurations for the given randomness factor is simply replaced by the results of the mitigation strategy (either a set of single value or a subset that is significantly smaller). The rest of the method remains unchanged. (line 3 in the Algorithm~\ref{alg:investigation})
    \item After enough configurations of non-investigated randomness factors (i.e., \textit{mitigation runs}) are searched through and enough runs of training and evaluation are performed, the partial standard deviations are averaged to produce the contributed standard deviation, and the partial means are aggregated (by taking their standard deviation) to produce the mitigated standard deviation. (lines 11-12 in the Algorithm~\ref{alg:investigation}) 
    \item The golden model standard deviation is calculated by simply performing training and evaluation multiple times with differently fixed configuration of all randomness factors. If enough overall runs are used, the golden model standard deviation can be replaced by simply taking the standard deviation over all the runs in the investigation. However, this may lead to incorrect results. (final loop; lines 13-17 in the Algorithm~\ref{alg:investigation})
    \item The importance score of the investigated factor is calculated as a fraction of the golden model standard deviation of the difference between contributed standard deviation and the mitigated standard deviation (to determine how much more the investigated factor contributes over all the mitigated ones). Any randomness factor with importance score over $0$ is considered significantly important, as such factors contribute the same amount of deviation as the combination of all the remaining factors. At the same time, the size of the importance value determines the overall importance of the model (i.e., factor with importance score of $0.6$ is more important than the ones with score of $0.1$). (the final check; lines 18-21 in the Algorithm~\ref{alg:investigation})
\end{enumerate}

\subsection{Illustration of the Method and its Results}
\label{appendix:investigation-full-illustration}

\begin{table*}[tbh]
\begin{center}
\begin{small}
\begin{sc}
\begin{tabular}{cc|ccccc|cc}
\toprule
                    &           & \multicolumn{5}{c|}{$\mathbb{IFC}_i$}                         &                                    &                                        \\
                    &           & $n_1$       & $n_2$       & ... & $n_{N-1}$     & $n_N$       &                                    &                                        \\ \hline
                    & $m_1$     & $r_{1,1}$   & $r_{1,2}$   & ... & $r_{1,N-1}$   & $r_{1,N}$   & $p\_mean_{1}$                      & $p\_std_{1}$                           \\
                    & $m_2$     & $r_{2,1}$   & $r_{2,2}$   & ... & $r_{2,N-1}$   & $r_{2,N}$   & $p\_mean_{2}$                      & $p\_std_{2}$                           \\
$\mathbb{MFC}_i$    & ...       & ...         & ...         & ... & ...           & ...         & ...                                & ...                                    \\
                    & $m_{M-1}$ & $r_{M-1,1}$ & $r_{M-1,2}$ & ... & $r_{M-1,N-1}$ & $r_{M-1,N}$ & $p\_mean_{M-1}$                    & $p\_std_{M-1}$                         \\
                    & $m_{M}$   & $r_{M,1}$   & $r_{M,2}$   & ... & $r_{M,N-1}$   & $r_{M,N}$   & $p\_mean_{M}$                      & $p\_std_{M}$                           \\ \hline
                    &           &             &             &     &               &             & $m\_std = $                        & $c\_std = $                            \\
                    &           &             &             &     &               &             & $std(p\_mean_{*})$                 & $\overline{p\_std_{*}}$                \\

\bottomrule
\end{tabular}
\end{sc}
\end{small}
\end{center}
\caption{The effects of a \textit{randomness factor} $i$ are determined by observing the variability in results over its \textit{configurations}, while mitigating the effects of other \textit{randomness factors}. The results are first grouped by the \textit{mitigated factor configurations} $m$ and a partial mean ($p\_mean_{m}$) and standard deviation ($p\_std_{m}$) is calculated. These values are then aggregated into \textit{contributed standard deviation} ($c\_std$), representing the effects of investigated \textit{randomness factor}, by calculating a mean over the $p\_std_{m}$, and into \textit{mitigated standard deviation} ($m\_std$), representing the remaining effects of mitigated \textit{randomness factors}, by calculating a standard deviation over $p\_mean_{m}$.}
\label{tab:investigation-example-formal}
\end{table*}

In this section, we provide the visualisation of the method in a form of table. In essence, when investigating the specific randomness factor, while mitigating the effects of other randomness factors, we fill in such table as illustrated in Table~\ref{tab:investigation-example-formal}. The columns represent the different configurations for the investigated factor. Observing how the performance changes across these columns, we can determine the effects of the randomness factors -- aggregating across these columns we obtain the partial mean $p\_mean$ and partial standard deviation $p\_std$. 

However, having only a single row would not deal with the interactions. Therefore we perform this investigation multiple times, each time with different randomly fixed combination of configurations for all the other, non-investigated randomness factors. Each such repeat of the investigation run represents a single row in the table, each with its own partial mean $p\_mean_{m}$ and partial standard deviation $p\_std_{m}$. 

To get the final \textit{contributed standard deviation} $c\_std$ for the investigated randomness factor, we aggregate over these different partial standard deviations ($c\_std = \overline{p\_std_{*}}$). In addition, to obtain the \textit{mitigated standard deviation} $m\_std$ we aggregate over the partial means ($m\_std = std(p\_mean_{*})$).

\subsection{Selecting Number of Investigation and Mitigation Runs}
\label{appendix:selecting-n-m}

When selecting the number of investigation ($N$) and the number of mitigation ($M$) runs, we need to find a balance between how well the effects of the factors are estimated and how well the interaction between the effects of different randomness factors are mitigated, and how much computational resources are required to get to this estimation and mitigation. An optimal solution is to use the lowest number of overall runs (that lead to lowest computational resources) after which the change in the results (the contributed/mitigated standard deviation or the normalised importance score) is under an acceptable threshold $\epsilon$. The value of this threshold $\epsilon$ depends on the setup of the experiment and the goal of our investigation, as in some cases higher change in the standard deviation may be acceptable, while in others we require a more strict setting.

In this section, we describe a simple method to search for this optimal point that can be used instead of the heuristics at the end of Section~\ref{sec:methodology} (which were a result of our analysis using the following method). The method is composed of following steps:
\begin{enumerate}
    \item The threshold of smallest acceptable change $\epsilon$, and the starting number of investigation runs $N$ are selected. The number of investigation runs should be sufficiently high from the start (following recommendations in Section~\ref{sec:methodology}) to make the search faster.
    \item A new mitigation run should be performed using a randomly selected configuration of the non-investigated randomness (or the number of investigation runs should be increased, running the new investigation runs for all the already performed mitigation runs).
    \item The new values of the relevant metrics (contributed standard deviation, mitigated standard deviation, or the importance score) should be determined and the difference to previous values calculated.
    \item If the observed change is lower than the threshold $\epsilon$ the current values of hyperparameters $N$ and $M$ represent the optimal point and should be used. Otherwise, continue to step 2 (increasing either the value of $N$ or $M$).
\end{enumerate}

In case our goal is to use the results of the investigation and the importance score for a more in-depth analysis and comparison across different factors, models, datasets or other experimental settings, the method should be repeated for every setting and the highest values of $N$ and $M$ should be used -- to guarantee that the comparison and analysis is done on the same number of overall runs and to not introduce any possible biases into the comparison. 

\section{Ablation Study: Reducing Number of Mitigation Runs and Test Data Size}
\label{appendix:ablation-study}

As mentioned in Section~\ref{sec:methodology}, there is a trade-off between feasibility (computations costs) of the investigation and precision (reliability) of the investigation results. This trade-off mainly depends on the number of mitigation runs (i.e., the number of configurations explored for the non-investigated randomness factors). To determine the optimal number of mitigation runs, we explore this trade-off using a modified version of the method described in Appendix~\ref{appendix:selecting-n-m}: we run the investigation for a larger number of mitigation runs (observing the behaviour even after the optimal point) and explore the effects of reducing \textbf{the number of mitigation runs ($M$)} and the \textbf{number of test samples} used for evaluation on the results and how well they estimate the overall contributed effects and how well the interactions are mitigated. We perform this ablation study for the Flan-T5 model on the SST2 dataset and report only specific interesting points.

As the baseline for this ablation study we work with the setting of using 100 mitigation runs (with 10 investigation runs) and 100\% of test samples. For the number of mitigation runs, we explore: 1) increasing the number significantly (to 500); 2) reducing the number to 10\% (10 mitigation runs). For the number of test samples, we explore reducing the set to: 1) 1 000 samples (which represents approximately 10\% of overall test samples); and 2) 500 samples (representing approximately 5\% of overall test samples). We also explore the combination of both reductions (in relevant cases). The results of this ablation study are available in Table~\ref{tab:ablation-reducing}.

\begin{table*}[tbh]
\begin{center}
\begin{small}
\begin{sc}
\begin{tabular}{@{}llcccccc@{}}

\toprule
Mitigation Runs &                          & \textbf{500}   & \textbf{100}   & \textbf{10}    & \textbf{100}   & \textbf{10}    & \textbf{10}    \\
Test Dataset Size    &                     & \textbf{100\%} & \textbf{100\%} & \textbf{100\%} & \textbf{$\sim$10\%}  & \textbf{$\sim$10\%}  & \textbf{$\sim$5\%}   \\ 
\% of Baseline Setting Data &              & \textbf{500\%} & \textbf{100\%} & \textbf{10\%}  & \textbf{$\sim$10\%}  & \textbf{$\sim$1\%}   & \textbf{$\sim$0.5\%} \\ \midrule

Golden          & \textit{F1 macro (\%)}   & 78.23          & 78.17          & 78.25          & 78.18          & 78.13          & 78.16          \\
Model           & \textit{F1 std}          & 2.31           & 2.24           & 2.09           & 2.50           & 2.35           & 2.97           \\ \midrule

Label           & \textit{F1 macro (\%)}   & 78.26          & 78.14          & 78.17          & 78.07          & 77.87          & 77.72          \\
Selection       & \textit{F1 std}          & 2.28           & 2.41           & 2.44           & 2.61           & 2.94           & 3.20           \\
                & \textit{Contributed std} & \textbf{2.073} & \textbf{2.167} & \textbf{2.135} & \textbf{2.204} & \textbf{2.174} & \textit{2.188} \\
                & \textit{Mitigated std}   & 0.797          & 0.904          & 0.946          & 1.278          & 1.806          & 2.193          \\ 
                & \textit{Importance}      & 0.55           & 0.56           & 0.57           & 0.37           & 0.16           & -0.00          \\ \midrule

Data            & \textit{F1 macro (\%)}   & 78.18          & 78.24          & 77.98          & 78.39          & 78.22          & 78.33          \\
Split           & \textit{F1 std}          & 2.29           & 2.30           & 2.30           & 2.55           & 2.59           & 2.85           \\
                & \textit{Contributed std} & \textbf{2.112} & \textbf{2.128} & \textbf{2.138} & \textbf{2.372} & \textbf{2.422} & \textbf{2.670}  \\
                & \textit{Mitigated std}   & 0.712          & 0.693          & 0.662          & 0.708          & 0.729          & 0.788          \\ 
                & \textit{Importance}      & 0.61           & 0.64           & 0.71           & 0.67           & 0.72           & 0.63           \\ \midrule

Data            & \textit{F1 macro (\%)}   & 78.14          & 78.28          & 77.29          & 78.22          & 77.10          & 76.82          \\
Order           & \textit{F1 std}          & 2.28           & 2.15           & 2.59           & 2.34           & 3.18           & 3.25           \\
                & \textit{Contributed std} & 0.846          & 0.869          & 0.982          & 0.902          & 1.095          & 1.149          \\
                & \textit{Mitigated std}   & 2.089          & 1.928          & 2.334          & 2.117          & 2.932          & 2.988          \\ 
                & \textit{Importance}      & -0.54          & -0.47          & -0.65          & -0.49          & -0.78          & -0.62          \\ \midrule

Sample          & \textit{F1 macro (\%)}   & 78.22          & 78.19          & 78.15          & 78.14          & 77.92          & 77.80          \\
Choice          & \textit{F1 std}          & 2.14           & 2.35           & 2.64           & 2.55           & 2.87           & 3.05           \\
                & \textit{Contributed std} & \textbf{2.138} & \textbf{2.123} & \textbf{2.361} & \textbf{2.152} & \textbf{2.337} & \textbf{2.325} \\
                & \textit{Mitigated std}   & 0.818          & 0.844          & 1.001          & 1.248          & 1.553          & 1.906          \\
                & \textit{Importance}      & 0.57           & 0.57           & 0.65           & 0.36           & 0.33           & 0.14           \\ 
\bottomrule
\end{tabular}
\end{sc}
\end{small}
\end{center}
\caption{The effects of changing the number of mitigation runs and the number of samples used for evaluation on the results of our proposed investigation method when applied to Flan-T5 model used with SST-2 dataset. The column with 100 mitigation runs and 100\% test data represents our baseline setting. With decreasing number of mitigation runs and the size of test data used, the \textit{mitigated standard deviation}, as well as the overall standard deviation increases, while the \textit{contributed standard deviation} stays approximately the same. This leads to lower precision of the results and change in the importance score of the different factors, and even can lead to incorrect results in extreme cases (Label Selection not being considered important when using $\sim$0.5\% of data as compared to our baseline setting). Even with $\sim$1\% of computation (combination of mitigation runs and test sample reduction) the findings can be considered sufficiently reliable in this setting.}
\label{tab:ablation-reducing}
\end{table*}

Compared to our baseline setting for the experiments (100 mitigation runs, with 100\% of test samples used), increasing the number of mitigation runs by 500\% does not lead to a significantly more reliable results. We can observe a slight change in overall standard deviation in the model (ranging from a change of $0.01$ to change of $0.21$). Similarly, the observed contributed standard deviation, as well as the mitigated standard deviation stays approximately the same (the change ranging from $0.005$ to $0.1$). In addition, the change in importance score is negligible for the different factors. All in all, we can conclude that increasing the number of mitigation runs any further does not make sense in regards to the reliability-feasibility trade-off.

On the other hand, reducing the number of mitigation runs and the number of test samples used for evaluation, we can observe more significant changes in the overall variance in the model and the importance score of the factors. We can observe a progressive increase in the overall golden model standard deviation (from $2.24$ up to $2.97$ in the most extreme setting). At the same time, we also observe significant increase in the mitigated standard deviation (going from as low as $0.904$ in the Label Selection randomness factor up to $2.193$ for the same factor in the most extreme setting), which can be expected as the number of mitigation runs governs the mitigation of non-investigated, interacting randomness factors in our method. Similarly, we can observe change in the importance score as well, with the importance of different factors being lower with lower number of mitigation runs (with the exception of Data Split randomness factor). In the most extreme setting (using 10 mitigation runs and 500 test samples, which represents $\sim$0.5\% of baselines setting data) we can even observe a change in the findings regarding the Label Selection randomness factor -- it becomes non-important as it is overshadowed by the mitigated randomness factors, with the importance score being slightly below the $0$ value. However, the less extreme setting, where $\sim$1\% of the baseline setting data is used (10 mitigation runs and 1000 samples), the results are still reliable enough (even though the importance score is lower in this case). In addition, the difference in importance score when using smaller amount of test samples is more significant than when using smaller number mitigation runs (i.e., $0.36$ and $0.33$ importance with 10\% test data while using 100 and 10 mitigation runs respectively, as compared to $0.57$ and $0.65$ when using full test data and 100 and 10 mitigation runs respectively). All in all, we can conclude that our proposed method is not as dependent on the number of mitigation runs and not as computationally expensive as can be expected, making it more easily usable on more computationally expensive settings (e.g., having large labelled datasets or using more computationally expensive models). At the same time, the importance score is dependent on the number of test samples used for evaluation, which needs to be taken into consideration when using it on setting such as in-context learning, where the inference itself is quite expensive.

Even when reducing the computation cost of the proposed method to $\sim$1\% of our baseline setting (reducing the number of mitigation runs to 10 and using only 1 000 test samples for evaluation) the findings can be considered sufficiently reliable. Therefore, if the precision of the results is not as paramount, the proposed method can be used even in this reduced setting (although one needs to be aware of the implications). To produce more precise results, and due to the significant computation costs of running the larger in-context learning models (LLaMA-2, Mistral and Zephyr), we have decided to run the investigation using 20 mitigation runs and 1 000 test samples (following the practice in related work~\cite{gao-etal-2021-making, chang-jia-2023-data, sclar2023quantifying, li-qiu-2023-finding, koksal-etal-2023-meal}). As such, the observed importance scores for different factors may be affected by this choice, but the findings regarding the importance should still hold. In addition, to keep the comparison between models as unbiased as possible, we use the same amount of test data for all the models and all the datasets and across all experiments.

Based on the observed behaviour, we can determine which factor affects the variability of the model results the most -- Data Split. For all the randomness factors, except for the Data Split, only the mitigated standard deviation increases when reducing the number of mitigation runs and/or the number of samples, while the contributed standard deviation stays approximately the same. However, for the Data Split randomness factor, the exact opposite happens (contributed std increases, while mitigated std stays the same). In essence, having more mitigation runs and/or using more test samples for evaluation leads to a significant mitigation of the variance from the data split randomness factor.

\section{Experimental Setup and Implementation Details}
\label{appendix:implementation-experimental-setting}

\begin{table*}[!tbh]
\begin{center}
\begin{small}
\tabcolsep=0.11cm
\begin{tabularx}{\textwidth}{@{}llp{0.35\linewidth}p{0.5\linewidth}@{}}
\toprule
\textbf{Dataset} & \textbf{ID} & \textbf{Verbaliser} & \textbf{Prompt Format}  \\ \midrule

SST-2   & A & \{Negative, Positive\} & Determine sentiment of the sentence using following options: 1) \textit{[Class 1]} 2) \textit{[Class 2]}. \newline \textit{[Input]} \newline \textit{[Output]}  \\ 
& B & Same as above & \textit{[Input]} Sentiment? \textit{[Output]}  \\ 
& C & Same as above & \textit{[Input]} Sentiment is \textit{[Output]}  \\ 
& D & \{terrible, great\} & \textit{[Input]} It was \textit{[Output]}  \\ \midrule

CoLA   & A & \{No, Yes\} & Determine grammatical acceptability of the sentence using following options: 1) \textit{[Class 1]} 2) \textit{[Class 2]}. \newline \textit{[Input]} \newline \textit{[Output]}  \\ 
& B & Same as above & \textit{[Input]} Grammatically acceptable? \textit{[Output]}  \\ 
& C & \{Yes, No\} & \textit{[Input]} Grammar problems? \textit{[Output]}  \\ 
& D & \{not acceptable, acceptable\} & \textit{[Input]} It is \textit{[Output]}  \\ \midrule

MRPC   & A & \{No, Yes\} & Determine whether the sentence pair is semantically equivalent using following options: 1) \textit{[Class 1]} 2) \textit{[Class 2]}. \newline \textit{[Input]} \newline \textit{[Output]}  \\ 
& B & Same as above & \textit{[Input]} Semantically equivalent sentences? \textit{[Output]}  \\ 
& C & \{Yes, No\} & \textit{[Input]} Semantically different sentences? \textit{[Output]}  \\ 
& D & \{not equivalent, equivalent\} & \textit{[Input]} Sentences are \textit{[Output]}  \\ \midrule

AG News   & A & \{World, Sports, Business, Science and Technology\} & Determine topic of the sentence using following options: 1) \textit{[Class 1]} 2) \textit{[Class 2]} ... N) \textit{[Class N]}. \newline \textit{[Input]} \newline \textit{[Output]}  \\ 
& B & Same as above & \textit{[Input]} Topic? \textit{[Output]}  \\ 
& C & Same as above & \textit{[Input]} Topic is \textit{[Output]}  \\ 
& D & Same as above & User: \textit{[Input]} This is about \textit{[Output]}  \\ \midrule

TREC   & A & \{Expression, Entity, Description, Human, Location, Number\} & Determine topic of the sentence using following options: 1) \textit{[Class 1]} 2) \textit{[Class 2]} ... N) \textit{[Class N]}. \newline \textit{[Input]} \newline \textit{[Output]}  \\ 
& B & Same as above & \textit{[Input]} Topic? \textit{[Output]}  \\ 
& C & Same as above & \textit{[Input]} Topic is \textit{[Output]}  \\ 
& D & Same as above & User: \textit{[Input]} This is about \textit{[Output]}  \\ \midrule

SNIPS   & A & \{Playlist, Weather, Event, Musing, Creative Work, Rate Book, Book Restaurant\} & Determine intent of the sentence using following options: 1) \textit{[Class 1]} 2) \textit{[Class 2]} ... N) \textit{[Class N]}. \newline \textit{[Input]} \newline \textit{[Output]}  \\ 
& B & Same as above & \textit{[Input]} Intent? \textit{[Output]}  \\ 
& C & Same as above & \textit{[Input]} Intent is \textit{[Output]}  \\ 
& D & Same as above & User: \textit{[Input]} User requested \textit{[Output]}  \\ \midrule

DB-Pedia   & A & \{Company, Educational Institution, Artist, Athlete, Office Holder, Transportation, Building, Natural Place, Village, Animal, Plant, Album, Film, Written Work\} & Determine topic of the sentence using following options: 1) \textit{[Class 1]} 2) \textit{[Class 2]} ... N) \textit{[Class N]}. \newline \textit{[Input]} \newline \textit{[Output]}  \\ 
& B & Same as above & \textit{[Input]} Topic? \textit{[Output]}  \\ 
& C & Same as above & \textit{[Input]} Topic is \textit{[Output]}  \\ 
& D & Same as above & User: \textit{[Input]} This is about \textit{[Output]}  \\
\bottomrule
\end{tabularx}
\end{small}
\end{center}
\caption{Prompt formats and verbalisers used for different datasets in the paper. The \textit{[Class 1-N]} are replaced with the names of the classes as defined by the verbaliser. The \textit{[Input]} is replaced by the sentence of the samples and the \textit{[Output]} is replaced with the name of class as defined by the verbaliser. The \textit{[Input]} and \textit{[Output]} are repeated for each in-context sample, while the final \textit{[Output]} is used to determine the predicted class. The same format is used for all the language models (Flan-T5, LLaMA-2-13B, Mistral-7B and Zephyr-7B).}
\label{tab:prompt-format}
\end{table*}

All the experiments in this paper are using English only datasets from the GLUE~\cite{wang-etal-2018-glue} benchmark suite and other publicly available datasets. The datasets from GLUE benchmark, SST2~\cite{dankers-titov-2022-recursive}, CoLA~\cite{warstadt-etal-2019-neural} and MRPC~\cite{dolan-brockett-2005-automatically}, are all binary classification datasets using only 2 classes. The remaining datasets represent a multi-class classification problems, with the AG News~\cite{zhang2015agnews} dataset consisting of 4 classes, TREC~\cite{voorhees2000trec} dataset consisting of 6 classes, SNIPS~\cite{coucke2018snips} dataset consisting of 7 classes and DB-Pedia~\cite{lehmann2015dbpedia} dataset consisting of 14 classes.

Based on the ablation study (included in Appendix~\ref{tab:ablation-reducing}), we use 10 investigation and 20 mitigation runs (resulting in overall 200 training and evaluation runs) for the in-context learning (Flan-T5, LLaMA-2, Mistral-7B and Zephyr-7B) and 100 mitigation runs (results in overall 1 000 training and evaluation runs) for the other approaches that use smaller models (BERT, RoBERTa). Following the practice from the related work (e.g.,~\cite{gao-etal-2021-making, chang-jia-2023-data, sclar2023quantifying, li-qiu-2023-finding, koksal-etal-2023-meal}) and the results of our ablation study, we evaluate each run using only 1 000 test samples (the selection is governed by the Label Selection randomness factor). The main reason is the computation cost of the inference for the large language models. To prevent introduction of any biases into the comparison, we use the same amount of test samples for the transfer learning and meta-learning as well (although we use larger number of runs results in larger distributions in those cases). These decisions represents the trade-off between feasibility/required computation costs to achieve the results and how well the effects of randomness factors are estimated and the interactions between them mitigated.

Besides the factors that we focus our investigation on (Label Selection, Data Split, Model Initialisation, Data Order and Sample Choice), we also focus on mitigating other factors that we call as \textbf{Model Randomness}. This group of factors encompasses the randomness originating from use of non-deterministic operations in the model (e.g., dropout or sampling in the in-context learning models that generate text) and from implementation level factors (e.g., the impact of different libraries, non-deterministic CUDA operations or using different GPU types). To mitigate these effects, we set CUDA to deterministic, use the same library versions and the same GPUs throughout the experiments (one exception are the meta-learning experiments which were done on a separate GPU), while also setting a specific random seed that governs the non-deterministic operations in the models during training and inference (this seed is explored using the mitigation runs, so each experiment explored 20 or 100 different sets of this non-determinism).

For the in-context learning models, we use the Flan-T5 base model\footnote{\url{https://huggingface.co/google/flan-t5-base}}, the LLaMA-2 13B instruction optimised model\footnote{\url{https://huggingface.co/meta-llama/Llama-2-13b-chat-hf}}, Mistral-7B instruct fine-tuned model\footnote{\url{https://huggingface.co/mistralai/Mistral-7B-Instruct-v0.1}} and Zephyr-7B instruct fine-tuned model\footnote{\url{https://huggingface.co/HuggingFaceH4/zephyr-7b-alpha}} (alpha version as it worked better on the classification tasks than the beta model, due to the beta model generating large quantities of text and multiple classes at the same time). The LLaMA-2, Mistral and Zephyr models are all used in the 4-bit quantised setting. All of these models are set to produce deterministic output, while the number of tokens they can generate is limited to 10. In the majority of the setting, we use 2 samples per class, which are randomly sampled from the train dataset. We use only 2 samples, as the Flan-T5 model falls apart and starts predicting a single class for every test sample when using larger number of samples. We perform only a basic prompt engineering for these models (exploring also optimal prompt formats from related research papers~\citep{li-qiu-2023-finding, gao-etal-2021-making, koksal-etal-2023-meal}, the prompt format recommended for the LLaMA-2 model, and taking inspiration from~\cite{sun2023pushing}), while also using the meta-tags that specify instruction for the models. The optimal prompt-format, as well as other formats used in the analyses, is illustrated in Tabled~\ref{tab:prompt-format}. In case the models produce multiple words that can be mapped to multiple classes (with the exception of specific prompts where some classes are subsets of each other), we treat the output as incorrect with the assumption the model is just hallucinating (although we noticed the Mistral and Zephyr models provide more detailed answers, especially on the SST2 dataset, which may lower their performance in this case).

For the fine-tuning models, BERT\footnote{\url{https://huggingface.co/bert-base-uncased}} and RoBERTa\footnote{\url{https://huggingface.co/roberta-base}}, we use the base version of the pre-trained models from HuggingFace~\cite{wolf2019huggingface}. Both models are trained in full (without freezing the pre-trained part) on all datasets using learning rate of 1e-5 for 5 epochs on binary and 10 epochs on multi-class dataset, using early stopping, AdamW optimiser with warmup for 10\% of the steps and batch size 8.

As the basis for the meta-learning approaches, we use the implementation released by the authors of the specific papers when possible, while the individual implementations are extended and modified to better work with our proposed method for investigation. In case of the Prototypical Networks, we directly use the code released by the authors\footnote{\url{https://github.com/jakesnell/prototypical-networks}}. In case of Model Agnostic Meta-Learning, we use the implementation from the Torchmeta library\footnote{\url{https://github.com/tristandeleu/pytorch-meta}}. In case of Reptile, we use our own implementation based on the code released for the approach\footnote{\url{https://github.com/openai/supervised-reptile}}. For meta-learning, we use the same base model across all the meta-learning approaches. This model is a simple fully-connected layer with 128 neurons and a final classification layer on top of the BERT base model. Each meta-learning approach is trained in a 2-way 5-shot learning setup. For evaluation, the meta-learning models are first adapted using a single set of examples in 2-way 15-shot setting (examples are chosen based on the sample choice randomness factor) and then evaluated on the whole test dataset.

All the hyperparameters for all the models are set using a separate hyperparameter optimisation for both fine-tuning and meta-learning (we run no hyperparameter optimisation for in-context learning) using the validation data selected from the 1 000 training samples. This hyperparameter optimisation is done in a two-level fashion. First, the optimisation is run using large differences in the hyperparameter values, to find the approximate set of hyperparameters that should provide good performance on the given dataset. In the second step, we explore the hyperparameter space around these approximate hyperparameters, to find the optimal set of parameters. However, it is important to note that the hyperparameter search is performed on a fixed set of labelled samples, chosen beforehand, and on a single split, which may affect the optimal set of hyperparameters and lead to sub-optimal hyperparameters, especially in meta-learning.

When choosing the hyperparameter values in the first level, we draw inspiration from related work, using the optimal parameters reported in papers that propose, or use these approaches (such as~\citep{dodge_fine-tuning_2020, mccoy-etal-2020-berts, mosbach_stability_2021, sellam_multiberts_2022}. However, we also search through additional hyperparameter values besides those reported in related works to better explore the parameter space and obtain as precise results from the investigation as possible.


\section{Validating the Proposed Investigation Method}
\label{appendix:investigation-strategy-comparison}

In this Appendix, we provide further information on how the proposed investigation method was validated. As there is no ground-truth of the effects of randomness to compare against, the validity of methods for investigating the effects of randomness can be evaluated only indirectly. In this paper, we perform such indirect evaluation/validation by:
\begin{enumerate}
    \item \textbf{Evaluating the properties and benefits of the proposed methods by comparing it to the existing ones.} As discussed in the main content of the paper, the benefits of the proposed methods are: 1) importance score that can be used for more in-depth analysis (relative ordering of randomness factors and comparison across models, datasets and other experimental settings), as opposed to determining the importance only in binary fashion from previous works (factor is or is not important); and 2) handling interactions between effects of randomness factors, which, when previously ignored or not being addressed sufficiently caused inconsistencies in findings. We discuss this validation further in Appendix~\ref{appendix:validation-comparison-to-baselines} (which we consider as the main validation of the method).

    \item \textbf{Exploring how the results and findings change as we change the overall number of runs}. The results and findings of the method are dependent on the choice of how many investigation and mitigation runs are used. We observe a trade-off between how well the results are estimated (higher number of investigation runs leads to better estimation) and the interactions mitigation (higher number of mitigation runs leads to better mitigation), and the computation costs required to achieve the results (increasing the number of runs increases the overall costs). We discuss this validation further in Appendix~\ref{appendix:validation-exploring-hyperparameters}. 

    \item \textbf{Applying the method to different settings (factors, models, datasets) and observing the consistency of its results and findings.} As the investigation method is designed to be general, it should be applied across different experimental settings without showing any problems (i.e., working out-of-the-box on multiple factors, models and datasets). We discuss this validation further in Appendix~\ref{appendix:validation-consistency}.
\end{enumerate}

\subsection{Additional Results: Validation of Method Through Comparison with Typical Investigation Strategies}
\label{appendix:validation-comparison-to-baselines}

\begin{table*}[!tbh]
\begin{center}
\scriptsize
\begin{sc}
\tabcolsep=0.15cm
\begin{tabular}{@{}lrrr rrr rrr@{}}

\toprule
               & \multicolumn{3}{c}{SST2}  & \multicolumn{3}{c}{CoLA} & \multicolumn{3}{c}{MRPC} \\
Flan-T5        & Random                  & Fixed                   & Interactions   & Random                  & Fixed                   & Interactions  & Random                  & Fixed                   & Interactions  \\ \midrule
Golden Model   & 2.244                   & 2.244                   & 2.244          & 3.811                   & 3.811                   & 3.811         & 1.328                   & 1.328                   & 1.328         \\
Label Select.  & (*) 2.517               & (*) 2.594               & (*) 2.128      & (*) 3.602               & (*) 2.804               & (*) 3.257     & \textbf{(*) 1.122}      & \textbf{(*) 1.189}      & 0.363     \\
Data Split     & (*) 2.362               & (*) 2.480               & (*) 2.167      & (*) 3.961               & (*) 1.990               & (*) 3.483     & (*) 1.503               & \textit{0.252}          & (*) 0.926     \\
Data Order     & \textbf{(*) 2.131}      & \textbf{(*) 3.014}      &     0.869      & \textbf{(*) 3.122}      & \textbf{(*) 4.172}      & 1.793         & \textbf{(*) 1.007}      & 0.289                   & 0.209     \\ 
Sample Choice  & (*) 2.370               & (*) 3.191               & (*) 2.123      & (*) 3.478               & \textit{1.203}          & (*) 3.138     & \textbf{(*) 1.277}      & \textbf{(*) 0.678}      & 0.348     \\
\midrule \midrule
Zephyr-7B      & Random                  & Fixed                   & Interactions   & Random                  & Fixed                   & Interactions  & Random                  & Fixed                   & Interactions  \\ \midrule
Golden Model   & 1.043                   & 1.043                   & 1.043          & 9.566                   & 9.566                   & 9.566         & 12.785                  & 12.785                  & 12.785        \\   
Label Select.  & (*) 1.122               & (*) 1.004               & (*) 0.863      & (*) 7.367               & \textit{2.529}          & (*) 5.806     & \textbf{(*) 11.968}     & \textbf{(*) 11.977}     & 5.109         \\
Data Split     & (*) 1.185               & \textit{0.402}          & (*) 0.664      & (*) 8.235               & (*) 8.001               & (*) 7.675     & \textbf{(*) 12.504}     & \textbf{(*) 6.660}      & 5.973         \\
Data Order     & \textbf{(*) 1.138}      & \textbf{(*) 0.957}      &     0.456      & \textbf{(*) 9.622}      & \textbf{(*) 7.028}      &     3.598     & (*) 11.211              & \textit{4.913}          & (*) 8.038     \\ 
Sample Choice  & (*) 1.052               & \textit{0.406}          & (*) 0.744      & (*) 10.069              & \textit{4.379}          & (*) 9.135     & \textbf{(*) 12.980}     & \textbf{(*) 12.239}     & 6.305         \\
\midrule \midrule
BERT           & Random             & Fixed              & Interactions   & Random                  & Fixed                   & Interactions  & Random                  & Fixed                   & Interactions            \\ \midrule
Golden Model   & 0.970              & 0.970              & 0.970          & 1.473                   & 1.473                   & 1.473         & 2.929                   & 2.929                   & 2.929                   \\
Label Select.  & (*) 1.096          & (*) 1.409          & (*) 0.927      & (*) 1.552               & (*) 1.103               & (*) 1.212     & (*) 2.760               & (*) 2.308               & (*) 2.168               \\
Data Split     & (*) 1.096          & (*) 1.272          & (*) 0.937      & (*) 1.409               & (*) 1.649               & (*) 1.250     & (*) 2.904               & (*) 3.132               & (*) 2.384               \\
Model Init.    & (*) 1.155          & (*) 1.197          & (*) 0.828      & (*) 1.523               & (*) 2.481               & (*) 1.059     & (*) 2.813               & (*) 1.997               & (*) 2.180               \\
Data Order     & (*) 1.082          & (*) 1.217          & (*) 0.852      & \textbf{(*) 1.639}      & \textbf{(*) 1.333}      & 1.086         & (*) 2.809               & (*) 3.971               & (*) 2.081               \\ 
\midrule \midrule
ProtoNets      & Random             & Fixed              & Interactions   & Random                  & Fixed                   & Interactions  & Random                  & Fixed                   & Interactions            \\ \midrule
Golden Model   & 0.940              & 0.940              & 0.940          & 2.111                   & 2.111                   & 2.111         & 1.789                   & 1.789                   & 1.789                   \\
Label Select.  & (*) 0.987          & (*) 0.857          & (*) 0.887      & (*) 2.109               & (*) 1.497               & (*) 1.924     & (*) 1.572               & \textit{0.451}          & (*) 1.448               \\
Data Split     & (*) 1.012          & (*) 1.041          & (*) 0.959      & (*) 2.188               & (*) 2.301               & (*) 2.010     & (*) 1.919               & (*) 1.006               & (*) 1.791               \\
Model Init.    & \textbf{(*) 0.892} & \textbf{(*) 0.845} &     0.658      & (*) 2.222               & (*) 1.582               & (*) 1.801     & \textbf{(*) 1.888}      & 0.610                   & 1.240               \\
Data Order     & (*) 0.929          & (*) 3.510          & (*) 3.233      & (*) 4.114               & \textit{0.439}          & (*) 3.346     & (*) 3.087               & (*) 6.590               & (*) 2.265               \\
Sample Choice  & \textbf{(*) 0.983} & \textbf{(*) 0.832} &     0.646      & (*) 2.163               & \textit{0.890}          & (*) 1.659     & \textbf{(*) 1.805}      & \textbf{(*) 1.271}      & 1.084               \\

\bottomrule

\noalign{\vskip 2mm}
\toprule
               & \multicolumn{3}{c}{AG News}  & \multicolumn{3}{c}{TREC} & \multicolumn{3}{c}{SNIPS} \\
Flan-T5        & Random                  & Fixed                   & Interactions   & Random                  & Fixed                   & Interactions  & Random                  & Fixed                   & Interactions  \\ \midrule
Golden Model   & 3.090                   & 3.090                   & 3.090          & 1.324                   & 1.324                   & 1.324         & 2.284                   & 2.284                   & 2.284         \\
Label Select.  & 0.980                   & 0.391                   & 0.556          & \textbf{(*) 1.502}      & \textbf{(*) 1.210}      & 0.683         & \textbf{(*) 3.081}      & \textbf{(*) 2.969}      & 1.581     \\
Data Split     & (*) 6.152               & \textit{0.594}          & (*) 3.777      & \textbf{(*) 1.247}      & \textbf{(*) 1.844}      & 0.892         & \textbf{(*) 2.855}      & \textbf{(*) 2.740}      & 1.602     \\
Data Order     & \textbf{(*) 2.962}      & \textbf{(*) 4.912}      &     0.686      & (*) 1.222               & (*) 1.005               & (*) 0.815     & (*) 2.040               & \textit{0.964}          & (*) 1.769     \\ 
Sample Choice  & (*) 1.982               & \textit{1.466}          & (*) 0.806      & \textbf{(*) 1.616}      & \textbf{(*) 1.344}      & 0.819         & \textbf{(*) 3.156}      & 0.970                   & 1.590     \\
\midrule \midrule
Zephyr-7B      & Random                  & Fixed                   & Interactions   & Random                  & Fixed                   & Interactions  & Random                  & Fixed                   & Interactions  \\ \midrule
Golden Model   & 2.066                   & 2.066                   & 2.066          & 3.884                   & 3.884                   & 3.884         & 4.132                   & 4.132                   & 4.132        \\   
Label Select.  & (*) 1.966               & \textit{1.008}          & (*) 1.460      & (*) 3.196               & (*) 3.988               & (*) 2.963     & (*) 2.687               & (*) 2.812               & (*) 2.935         \\
Data Split     & (*) 2.141               & (*) 2.452               & (*) 1.817      & (*) 3.554               & (*) 2.115               & (*) 3.221     & (*) 4.006               & (*) 3.580               & (*) 3.052         \\
Data Order     & \textbf{(*) 1.859}      & \textbf{(*) 2.243}      &     0.919      & \textbf{(*) 4.037}      & \textbf{(*) 3.990}      &     1.925     & (*) 3.838               & \textit{1.012}          & (*) 3.007     \\ 
Sample Choice  & (*) 2.358               & \textit{0.884}          & (*) 1.874      & (*) 4.021               & (*) 4.696               & (*) 3.279     & (*) 3.975               & \textit{1.311}          & (*) 3.331         \\
\midrule \midrule
BERT           & Random             & Fixed              & Interactions   & Random                  & Fixed                   & Interactions  & Random                  & Fixed                   & Interactions            \\ \midrule
Golden Model   & 1.239              & 1.239              & 1.239          & 1.667                   & 1.667                   & 1.667         & 0.486                   & 0.486                   & 0.486                   \\
Label Select.  & (*) 1.202          & (*) 0.923          & (*) 0.979      & (*) 1.600               & (*) 1.513               & (*) 1.348     & (*) 0.559               & (*) 0.405               & (*) 0.308               \\
Data Split     & (*) 1.462          & (*) 1.365          & (*) 1.164      & (*) 1.502               & (*) 1.513               & (*) 1.568     & (*) 0.401               & (*) 0.426               & (*) 0.294               \\
Model Init.    & \textbf{(*) 1.142} & \textbf{(*) 1.047} &     0.693      & \textbf{(*) 1.926}      & \textbf{(*) 1.108}      &     0.939     & \textbf{(*) 0.479}      & \textbf{(*) 0.635}      &     0.121               \\
Data Order     & \textbf{(*) 1.335} & \textbf{(*) 0.714} &     0.686      & \textbf{(*) 1.666}      & \textbf{(*) 1.391}      &     1.019     & \textbf{(*) 0.471}      &     0.173               &     0.103               \\ 

\bottomrule
\end{tabular}
\end{sc}
\end{center}
\caption{Comparison of different investigation strategies for the Flan-T5, Zephyr-7B and BERT fine-tuning on the binary datasets (SST2, CoLA and MRPC) and the multi-class datasets (AG News, TREC and SNIPS). Comparison for the DB Pedia dataset is not included as Flan-T5 model shows poor performance on this particular dataset. The `Random` strategy simply repeats the training and evaluation multiple times without any constraints. In the `Fixed` strategy, the \textit{randomness factor configuration} is kept fixed to a single state during investigation. We compare these investigation strategies with our proposed method. We run each investigation strategy the same number of times (number of runs is governed by our method). Our method (`Interactions`) takes the interactions into consideration. Factors considered important for different strategies are denoted using the (*) symbol. We observe that interactions between factors may cause some factors to have their importance overestimated (denoted in \textbf{bold}) or underestimated (denoted in \textit{italics}).}
\label{tab:comparison-baseline-binary}
\end{table*}

To showcase the impact of interactions between \textit{randomness factors} on the investigation of the effects of different randomness factors, and to showcase the properties and benefits of our proposed method, we provide a comparison between the typical investigation strategies from related work and our proposed method:
\begin{itemize}
    \item \textbf{Random} -- investigation strategy without any constraints on the \textit{randomness factor configurations}. For each training and evaluation run of the model, all the \textit{randomness factors} are varied, while only the impact of a specific factor is observed. For example, each training and evaluation is done on a different set of training and testing data, with different order in training and with different random model initialisation, regardless which randomness factor is investigated. This represents the typical investigation process when considering only the random seed randomness factor. This investigation strategy does not consider any impact of interactions between randomness factors. As there is no change in how the individual randomness factor is investigated, we expect most skewed results from this investigation strategy, with each randomness factors showing approximately similar effects.

    \item \textbf{Fixed} -- investigation strategy where the interactions are addressed by fixing the non-investigated \textit{randomness factors} to a single \textit{randomness factor configuration}. For example, each training and evaluation is done on the same set of training and testing data, with the data in the same order, but each time with different random initialisation of the model. However, as only a single randomness factor configuration is used for the non-investigated randomness factors, the effects of the investigated randomness factor may still be affected by the interactions (due to the randomly chosen point in the randomness factor configuration state space). Therefore we expect the results to represent the effects of different \textit{randomness factors} more accurately, but still can under-estimate or over-estimate some effects due to the still present randomness in the investigation.

    \item \textbf{Interactions (Our)} -- the investigation method proposed in this paper. In essence can be viewed as repeating the `Fixed` investigation strategy multiple times, each time with differently fixed \textit{randomness factor configurations}, and averaging over these repeats. 
\end{itemize}

To prevent introduction of any biases into the comparisons between the strategies, we perform same number of training and evaluation runs for each method. For each strategy, we repeat the training and evaluation 1 000 times (or 200 times, as governed by the number of runs in our proposed method). The full results are presented in Table~\ref{tab:comparison-baseline-binary} (except for DB Pedia dataset where the Flan-T5 model does not work well). We focus on 2 main aspects in the comparison: 1) determining importance of the factors; 2) how interactions affect the findings.

\paragraph{Determining importance of the factors.} As the \textit{Random} and \textit{Fixed} strategies results only in a single score (deviation in the results), we consider the factor to be important when it contributes at least 50\% of the golden model standard deviation. As such, the importance of the randomness factors can be determined only in a binary fashion (factor is or is not important). Such setting allows only for a limited analysis (only relative ordering of factors based on the deviation withing the same setup) and cannot be easily used to compare the importance across different models. On the other hand, our proposed method provides an importance score that can be used for more in-depth analysis, such as the relative ordering of randomness factors based on their importance, or comparison across models, datasets and experimental settings (as the importance score is normalised with the overall deviation in the results from the golden model). This benefit can be illustrated using following example -- using Table~\ref{tab:comparison-baseline-binary} and the \textit{Random}/\textit{Fixed} strategy, we cannot say with good conscience that the Sample Choice is more important for the Flan-T5 model than for the Zephyr-7B model based only on their standard deviation ($2.370$ vs. $1.052$ using \textit{Random}; $3.191$ vs. $0.406$ using \textit{Fixed}) as the overall deviation in results is higher, but can be done so using our method (importance score from Figure~\ref{fig:rf-importance-full} or Table~\ref{tab:flan-t5-results-full} and \ref{tab:zephyr-results-full} of $0.57$ vs. $0.39$) as the score is normalised. Or similarly for Data order ($2.131$ vs. $1.138$ for \textit{Random}; $3.014$ vs. $0.957$ for \textit{Fixed}; $-0.47$ vs. $-0.44$ for our method) -– using the importance score we see the importance of Data Order is similar (slightly higher for Zephyr-7B) for both models, while other investigation strategies show a large difference (higher importance for Flan-T5).

\paragraph{Handling interactions.} The existing strategies either ignore the interactions completely (\textit{Random}) or do not addresses the sufficiently (i.e., in a way that strongly depends on randomness in the \textit{Fixed} strategy). As such, the baseline strategies often lead to incorrect attribution of the effects of different factors, either due to overestimating the impact of non-important randomness factors, or underestimating the impact of important factors. For example, in the case of Flan-T5 in-context learning, these investigation strategies indicate that all the \textit{randomness factors} are equally important (as they contribute similar deviation to the golden model), which is not the case when the interactions are taken into consideration (when interactions are considered, the impact of data order falls off). In case of the \textit{Random} strategy, this behaviour stems from the strategy consistently leading to the same overall deviation/importance for all the investigated randomness factors (which is similar to the deviation of the Golden Model). Even though using the \textit{Fixed} investigation strategy produces more reliable results (which are more distributed and handle the interactions to a certain extent), it is still affected by the randomness caused by the choice of the single \textit{randomness factor configurations} for the non-investigated factors. The results still show both overestimation and underestimation of effects for the \textit{randomness factors}. On the other hand, our method is specifically designed to handle the interactions using the mitigation runs. Handling the interactions this way, we observe that the finding that the long believed sensitivity of in-context learning to Data Order is actually a sensitivity to Sample Choice (and potentially the choice of prompt format) when choosing samples in a more sophisticated manner, holds even when choosing samples at random.

All in all, our proposed method provides 2 significant benefits over the baseline strategies, which indirectly validates its use: 1) allowing for more in-depth analysis and comparison across different factors, models, datasets and experimental setups that leads to actionable findings and read-to-apply take-away messages and suggestions (described in experimental results in Sections~\ref{section:rf-importance}, \ref{section:classes-and-shots} and \ref{section:prompt-format}, such as increasing the number of shots for in-context learning reduces the importance of sample choice, but does not affect the importance of sample order); and 2) handling of interactions that leads to more consistent results.

\subsection{Additional Results: Validation of Method by Exploring the Changes Due to Different Number of Runs}
\label{appendix:validation-exploring-hyperparameters}

The results and findings from investigation are heavily dependent on the overall number of runs. As opposed to the baseline strategies, our proposed method introduces another parameter, number of mitigation runs, to handle the interactions. We provide results from exploring how changing the number of investigation and mitigation runs affect the results and findings (i.e., how well the effects are estimated and the interactions mitigated) in Appendix~\ref{appendix:selecting-n-m} and Appendix~\ref{appendix:ablation-study}, while in this section, we provide a summary of relevant results.

The effects of randomness factors can be estimated using a relatively low number of investigation runs (around $6$ to $8$). Increasing the number of investigation runs further does not lead to considerable changes in the estimated effects (the contributed standard deviation changes only in second decimal place).

On the other hand, increasing the number of mitigation runs has a larger impact on the overall results and findings (and the different metrics we use), as it represents the main avenue for mitigating the interactions. Any change to the number of mitigation runs changes all the metrics (contributed std, mitigated std, and the importance score). In addition, the number of mitigation runs also depends on the approach, model and dataset used. As such, it is important to find the optimal point, where the interactions are sufficiently handled without requiring extensive computation costs. To find this optimal point, we provide heuristics and a simple search method in Appendix~\ref{appendix:selecting-n-m}. However, the overall number of required mitigation runs is still relatively low -- in our experiments, we observed that using 20 mitigation runs provides sufficient mitigation of interactions and estimation of the overall effects.

Finally, we observed that the number of test samples used for evaluation is the most important factor influencing the estimation of the effects. In our experiments, we observed that using 1000 test samples for evaluation provides a good trade-off between the feasibility of larger scale experiments (due to computation costs) and the validity of the results. 

\subsection{Additional Results: Validation of Method by Observing Consistency of Results and Findings Across Different Settings}
\label{appendix:validation-consistency}

The proposed investigation method is designed to not be dependent on any specific experimental setup, so that it can be used across any randomness factors, model, dataset or other systematic changes (e.g., number of samples, or prompt formats). To validate this property of the method, we apply it across various settings and observe how consistent are the results and findings (the full results presented throughout the paper, such as in Tables~\ref{tab:flan-t5-results-full}-\ref{tab:reptile-results-full}, or Figures~\ref{fig:rf-importance-full}, \ref{fig:rf-importance-shots}, \ref{fig:rf-importance-format}, or in Appendix~\ref{appendix:results}):
\begin{itemize}
    \item Different randomness factors that require different configuration setup for investigation (e.g., different choice of data, order of samples, initialisation, etc.), but also different setup for their mitigation. As discussed in Appendix~\ref{appendix:investigation-highlevel}, the mitigation can be done on group level (effectively mitigating multiple randomness factors at the same time), while also allowing for further extensions (such as using different mitigation strategies).
    \item Different approaches, namely in-context learning, fine-tuning and meta-learning, and different models in these approaches. Although each approach works differently (e.g., fine-tuning using optimisation, while in-context learning uses only inference with prompts), the proposed method works with any such approach. The only limitation is that the models and approaches used have an option to allow for deterministic behaviour. Without this option, the method can still be applied but may produce inconsistent and non-reproducible results and findings (i.e., the importance score in such cases is affected by the non-determinism of the model and so cannot be trusted fully). In addition, we apply the proposed method to models that lead to different performance and show different overall deviation in the results. In all the cases, the produced importance score can be used for the analysis and comparison, even in cases when the impact of the randomness factor is significant (e.g., Prototypical Networks on the SST2 datasets with Data Order randomness factor, where we observe a significant drop in performance and increase in overall deviation as opposed to the golden model -- in which case the method correctly identifies this factor to be significantly important, leading to importance score of $0.92$)
    \item Datasets with different characteristics and different experimental setups, such as different number of classes, samples, different prompt formats.
\end{itemize}

In all of these cases, the proposed method produces consistent results and findings without any obvious shortcomings. Although the baseline strategies can also be applied across all the settings, they often lead to inconsistent results due to the mishandling of interactions (i.e., \textit{Random} consistently leads to results similar to Golden Model for all the randomness factors, while using \textit{Fixed} strategy the importance of different factors changes quite often across different models, approaches, datasets and experimental settings).

\section{Additional Results from Investigation}
\label{appendix:results}

\begin{figure*}[!tbh]
    \centering
    \includegraphics[width=.75\linewidth]{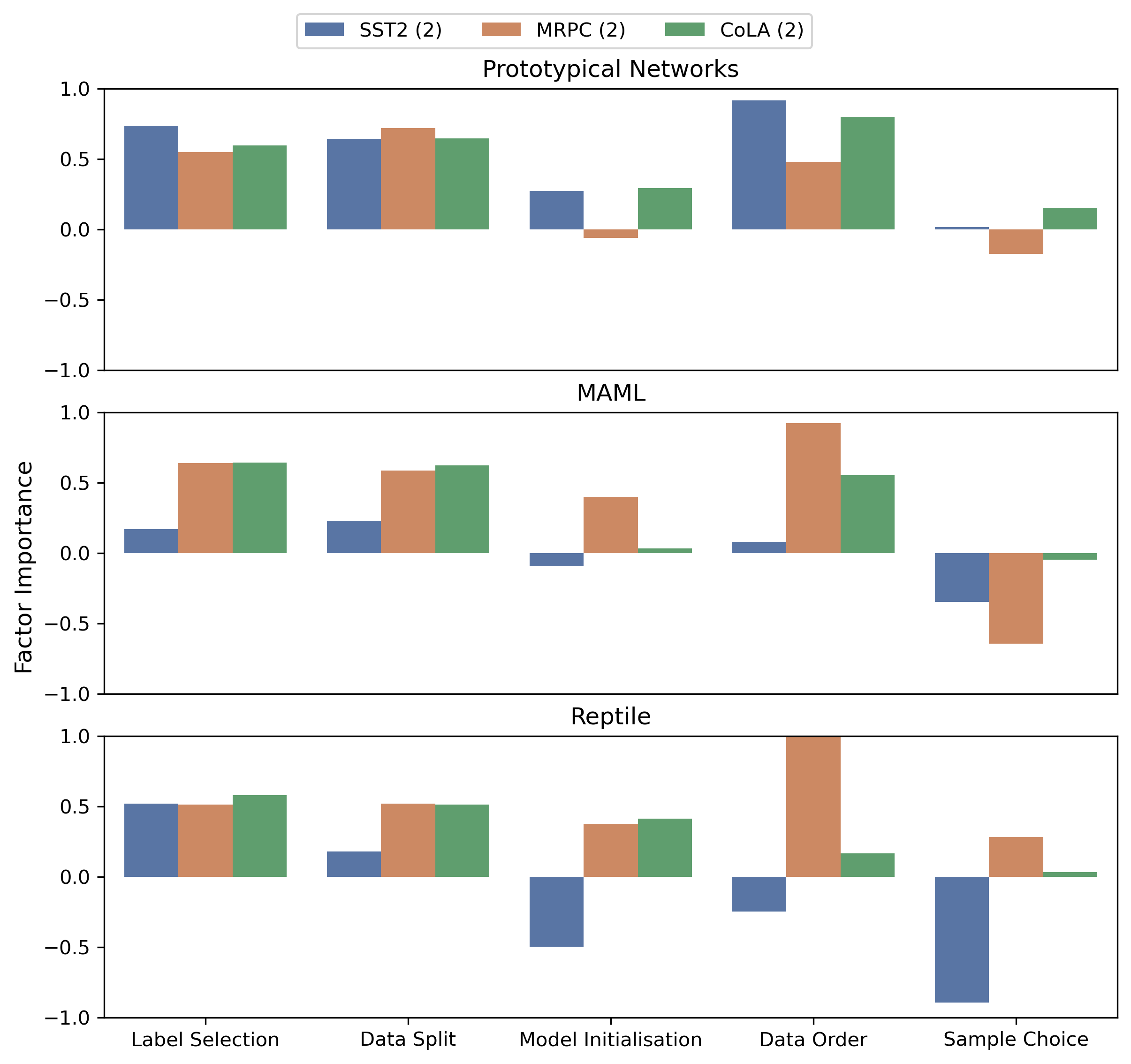}
    \caption{Importance of the investigated randomness factors for the meta-learning approaches on binary datasets, while taking the interactions between factors into consideration. The legend indicates number of classes for each datasets. We can observe consistent importance of the majority of the factors, with the exception of the Sample Choice and Model Initialisation factors. At the same time, the Data Order randomness factors appears to be the most important one for all the approaches.}
    \label{fig:rf-importance-metal}
\end{figure*}

\begin{figure*}[!tbh]
    \centering
    \includegraphics[width=1\linewidth]{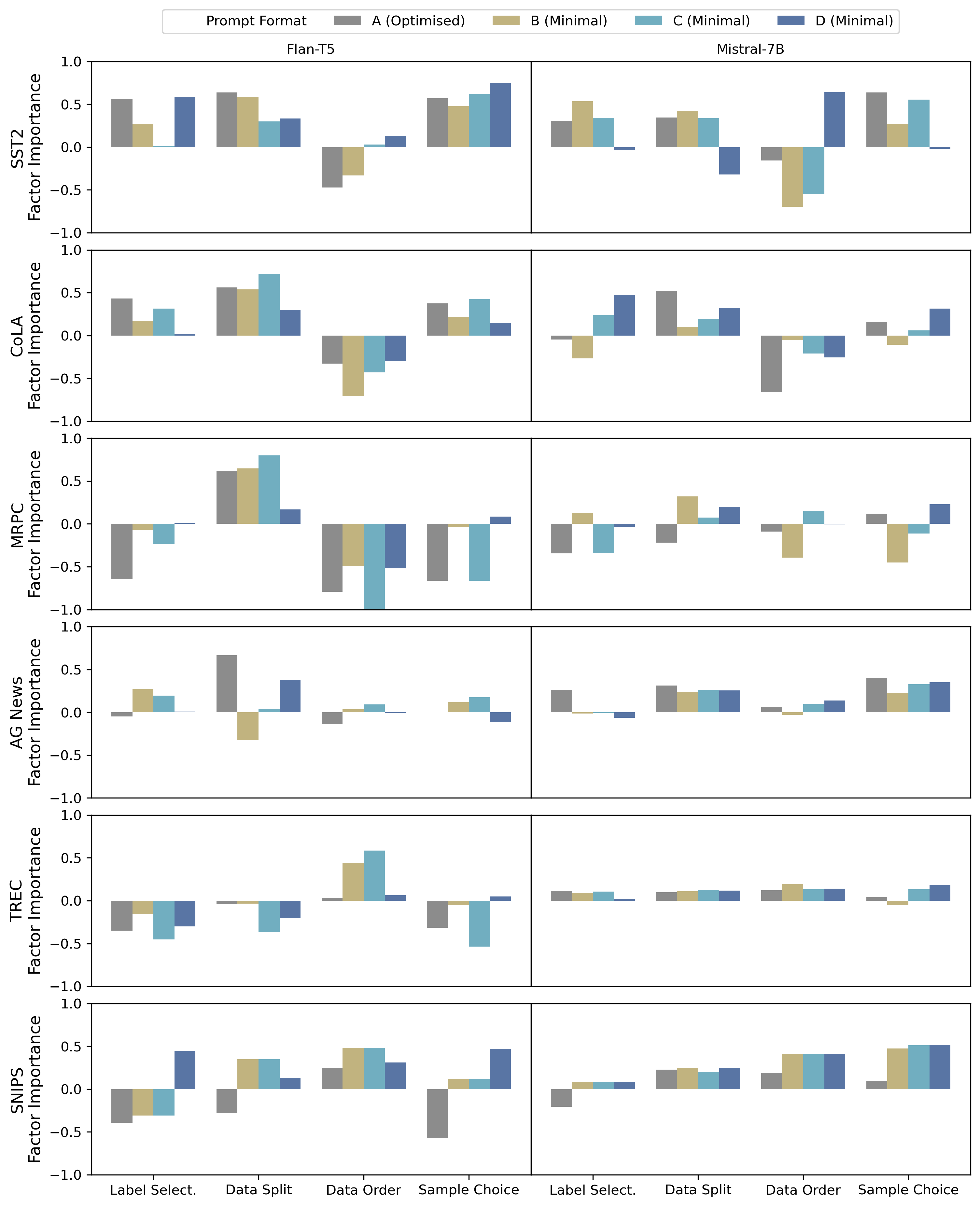}
    \caption{Effect of different prompt formats on the importance of randomness factors for in-context learning. The choice of format has significant effect on the importance of different factors, with the minimal formats often leading to higher importance. At the same time, the larger, more optimised models, show lower sensitivity to prompt format.}
    \label{fig:format-results-all-datasets}
\end{figure*}

In this Appendix, we provide additional results from the investigation experiments. This includes the investigation of randomness factor importance for the meta-learning approaches on the binary datasets (Appendix~\ref{appendix:metal-results}), the full results from investigating the impact of prompt format on the importance across all datasets (Appendix~\ref{appendix:format-results}, and the full results from the main investigation in a form of tables in order to present the performance and the deviation of different models and randomness factors (Appendix~\ref{appendix:table-results}).

\subsection{Additional Results: Meta-Learning Randomness Factor Importance on Binary Datasets}
\label{appendix:metal-results}
In this Appendix, we include the results of the randomness factor importance investigation for the meta-learning approaches on the binary datasets. The results are presented in Figure~\ref{fig:rf-importance-metal}.

For the majority of the approaches and the investigated datasets, the \textbf{Data Order randomness factor is the most important}, with the factors achieving importance score of 1.0 in some cases, which represents the situation, when the factor contributes all the deviation in the model. Even though this importance is due to the factor actually leading to significantly lower performance and significantly higher overall deviation when set to only specific subset, this only reinforces the finding that the Data Order factor is the most important.

In addition, we observe \textbf{a consistent importance of the Data Split and Label Selection randomness factors} for the meta-learning approaches across all the binary datasets. This follows the findings of transfer learning, which also performs optimisation/training and is not only composed of inference (as is the case with in-context learning). As such, we can conclude that the way the data is split and which samples are considered labelled has a significant impact on the approaches that require training. One possible reason is that the different splits and data labelling lead to different data distribution which severely affects the training.

Finally, the \textbf{Model Initialisation and Sample Choice (and task choice) randomness factors} do not show consistent importance across the meta-learning approaches and the datasets. However, the finding regarding Sample Choice may be due to the binary setting and may be different when using the meta-learning approaches in the true few-shot setting (i.e., using them to adapt to previously unseen classes and tasks).

\subsection{Additional Results: Impact of Prompt Format For All Datasets}
\label{appendix:format-results}

This Appendix contains the full results from investigating the impact of the prompt format on the effects of different randomness factors and their importance. The results for the Flan-T5 and Mistral-7B model across all the datasets are included in Figure~\ref{fig:format-results-all-datasets}. 

As already discussed in Section~\ref{section:prompt-format}, the format of the prompt used can have significant impact on the importance of different randomness factors. Using the minimal formats, we observe significant changes in the importance of different randomness factors, with them being not considered significantly important when using one format (e.g., Data Order on SST2 dataset using format B) and at the same time significantly important when using different format (e.g., Data Order on SST2 dataset using format D). 

In addition, the large language models are more robust to this change of prompt format. This finding is more evident on the multi-class datasets, where in comparison to Flan-T5 model, the importance score of the Mistral-7B remains more or less constant, while the importance score of Flan-T5 model oscillates significantly. On the binary datasets, the larger model is not as robust, but still the changes to the importance score are less significant than in the Flan-T5 model. Analysing the predictions further, we observe that the larger model provides more in-depth answers on the binary datasets (e.g., not providing only an answer but also an explanation for the answer, for example generating "positive (in a negative context)" instead of "positive", or often predicting neutral sentiment on the SST2 dataset, which is considered as incorrect answer), which may lead to the significant changes in the importance of the different randomness factors. 

\textbf{These findings only further highlights the importance of prompt-tuning as the format has significant impact on the words generated and therefore also the assigned classes and the importance scores of the different randomness factors.}

\subsection{Additional Results: Investigation Results in Table Form}
\label{appendix:table-results}

This Appendix contains the full results from the main investigation of the importance for the effects of different randomness factors in this work (which were included as Figure~\ref{fig:rf-importance-full}), in a form of tables with all the values included (performance, deviation, contributed deviation, mitigated deviation and importance for each investigated randomness factor). We believe that including these results allows for more in-depth analysis, exploration of the results and its further extension. In addition to the results, we provide a brief summary overview based on these results, which main not necessarily be connected only to the importance for different factors, but instead to the overall stability of the models and their ability to perform the different tasks.

The results are included as follows:
\begin{itemize}
    \item Flan-T5 results for all datasets (with exception of DB-Pedia) in Table~\ref{tab:flan-t5-results-full}
    \item LLaMA-2 results for all datasets in Table~\ref{tab:llama2-results-full}
    \item Mistral-7B results for all datasets in Table~\ref{tab:mistral-results-full}
    \item Zephyr-7B results for all datasets in Table~\ref{tab:zephyr-results-full}
    \item BERT results for all datasets in Table~\ref{tab:bert-results-full}
    \item RoBERTa results for all datasets in Table~\ref{tab:roberta-results-full}
    \item Prototypical Networks results for all binary datasets in Table~\ref{tab:protonet-results-full}
    \item MAML results for all binary datasets in Table~\ref{tab:maml-results-full}
    \item Reptile results for all binary datasets in Table~\ref{tab:reptile-results-full}
\end{itemize}

Based on these results, we can determine the overall stability of the different models. Specifically, we can observe the smaller in-context learning model (Flan-T5) shows better stability than the larger ones (LLaMA-2, Mistral-7B and Zephyr-7B), leading to significantly lower overall deviation across majority of the datasets. At the same time, we can observe that with increasing number of predicted classes, the performance of the Flan-T5 model drops significantly (from $83.85$ F1 on AG News dataset with 4 classes to $44.25$ on the SNIPS dataset with 7 classes), while retaining its stability (the overall deviation staying approximately the same with $3.09$ on AG News and $2.28$ on SNIPS). On the other hand, the larger language models achieve similar performance, but different stability, across the majority of the investigated datasets regardless of the number of predicted classes. The significant increase of performance and stability in case of the DB-Pedia dataset and SNIPS dataset (to a certain extent), may point to the fact that the models may have been trained on these datasets and so the results and findings on them may be biased -- we discuss this as a limitation based on the recently observed large language model validation crisis~\cite{li2023task}.

The fine-tuning approaches appear to be the most stable and best performing approaches in our investigation, leading to F1 score as high as $98\%$ and overall deviation as low as $0.36$. Surprisingly, the performance on the multi-class datasets is higher than on the binary datasets, which may indicate the overall ``hardness'' of the different datasets we use in this work, or point to specific problems in the binary datasets (such as the single word sentences without any sentiment in the SST2 dataset).

Finally, the meta-learning approaches appear to be significantly dataset dependent, with the overall performance and the overall deviation changing significantly across different binary datasets. One possibility for this is their significant sensitivity to the setup of the hyperparameters, with the performance and deviation changing significantly with even small changes in the hyperparameter setup, which we observed when trialling the meta-learning models on the multi-class datasets as well.

\begin{table*}[tbh]
\begin{center}
\begin{small}
\begin{sc}
\begin{tabular}{@{}llcccccc@{}}
\toprule
Flan-T5 &       & SST2  & CoLA  & MRPC  & AG News       & TREC  & SNIPS \\ \midrule 
Golden  & \textit{F1 Macro (\%)} & $78.17$       & $40.71$       & $70.70$       & $83.85$       & $61.87$       & $44.25$ \\ 
Model   & \textit{F1 Std}       & $2.24$        & $3.81$        & $1.32$        & $3.09$        & $1.32$        & $2.28$ \\ \midrule 

Label   & \textit{F1 Macro (\%)}        & $78.14$       & $40.65$       & $70.58$       & $84.31$       & $61.95$       & $43.80$ \\ 
Selection       & \textit{F1 Std}       & $2.41$        & $3.70$        & $1.27$        & $0.92$        & $1.37$        & $2.95$ \\ 
        & \textit{Contributed Std}      & $2.167$       & $3.257$       & $0.363$       & $0.556$       & $0.683$       & $1.581$ \\ 
        & \textit{Mitigated Std}        & $0.904$       & $1.610$       & $1.210$       & $0.711$       & $1.147$       & $2.476$ \\ 
        & \textit{Importance}   & $0.56$        & $0.43$        & $-0.64$       & $-0.05$       & $-0.35$       & $-0.39$ \\ \midrule 

Data    & \textit{F1 Macro (\%)}        & $78.24$       & $41.07$       & $70.78$       & $83.62$       & $61.73$       & $44.06$ \\ 
Split   & \textit{F1 Std}       & $2.30$        & $3.81$        & $0.94$        & $5.22$        & $1.36$        & $2.85$ \\ 
        & \textit{Contributed Std}      & $2.128$       & $3.483$       & $0.926$       & $3.777$       & $0.892$       & $1.602$ \\ 
        & \textit{Mitigated Std}        & $0.693$       & $1.344$       & $0.119$       & $1.717$       & $0.943$       & $2.244$ \\ 
        & \textit{Importance}   & $0.64$        & $0.56$        & $0.61$        & $0.67$        & $-0.04$       & $-0.28$ \\ \midrule 

Data    & \textit{F1 Macro (\%)}        & $78.28$       & $40.61$       & $70.60$       & $83.96$       & $62.11$       & $44.48$ \\ 
Order   & \textit{F1 Std}       & $2.15$        & $3.61$        & $1.27$        & $1.96$        & $1.14$        & $2.18$ \\ 
        & \textit{Contributed Std}      & $0.869$       & $1.793$       & $0.209$       & $0.686$       & $0.815$       & $1.769$ \\ 
        & \textit{Mitigated Std}        & $1.928$       & $3.044$       & $1.254$       & $1.115$       & $0.771$       & $1.197$ \\ 
        & \textit{Importance}   & $-0.47$       & $-0.33$       & $-0.79$       & $-0.14$       & $0.03$        & $0.25$ \\ \midrule 

Sample  & \textit{F1 Macro (\%)}        & $78.19$       & $40.68$       & $70.58$       & $83.87$       & $61.82$       & $43.77$ \\ 
Choice  & \textit{F1 Std}       & $2.35$        & $3.66$        & $1.27$        & $1.91$        & $1.51$        & $3.33$ \\ 
        & \textit{Contributed Std}      & $2.123$       & $3.138$       & $0.348$       & $0.806$       & $0.819$       & $1.590$ \\ 
        & \textit{Mitigated Std}        & $0.844$       & $1.711$       & $1.222$       & $0.786$       & $1.235$       & $2.897$ \\ 
        & \textit{Importance}   & $0.57$        & $0.37$        & $-0.66$       & $0.01$        & $-0.31$       & $-0.57$ \\
\bottomrule
\end{tabular}
\end{sc}
\end{small}
\end{center}
\caption{Results from investigating the importance for the effects of different randomness factors for the in-context learning using the Flan-T5 model across all datasets the model work correctly on.}
\label{tab:flan-t5-results-full}
\end{table*}

\begin{table*}[tbh]
\begin{center}
\begin{small}
\begin{sc}
\begin{tabular}{@{}llccccccc@{}}
\toprule
LLaMA-2-13B     &       & SST2  & CoLA  & MRPC  & AG News       & TREC  & SNIPS & DB-Pedia \\ \midrule 
Golden  & \textit{F1 Macro (\%)} & $90.48$       & $67.58$       & $58.84$       & $44.88$       & $39.85$       & $59.18$       & $62.34$ \\ 
Model   & \textit{F1 Std}       & $2.87$        & $4.12$        & $4.70$        & $5.51$        & $4.10$        & $5.82$        & $4.56$ \\ \midrule 

Label   & \textit{F1 Macro (\%)}        & $90.23$       & $66.35$       & $59.47$       & $45.67$       & $39.76$       & $59.27$       & $62.50$ \\ 
Selection       & \textit{F1 Std}       & $2.50$        & $4.10$        & $4.30$        & $4.98$        & $4.58$        & $5.35$        & $4.33$ \\ 
        & \textit{Contributed Std}      & $2.191$       & $3.470$       & $2.856$       & $4.077$       & $3.559$       & $3.118$       & $2.729$ \\ 
        & \textit{Mitigated Std}        & $1.036$       & $1.977$       & $3.065$       & $2.420$       & $2.602$       & $4.076$       & $3.248$ \\ 
        & \textit{Importance}   & $0.40$        & $0.36$        & $-0.04$       & $0.30$        & $0.23$        & $-0.16$       & $-0.11$ \\ \midrule 

Data    & \textit{F1 Macro (\%)}        & $90.16$       & $65.88$       & $58.51$       & $46.01$       & $39.41$       & $59.19$       & $61.72$ \\ 
Split   & \textit{F1 Std}       & $3.03$        & $3.73$        & $5.07$        & $5.90$        & $3.89$        & $4.52$        & $4.45$ \\ 
        & \textit{Contributed Std}      & $2.374$       & $2.853$       & $3.924$       & $4.312$       & $2.601$       & $3.707$       & $2.439$ \\ 
        & \textit{Mitigated Std}        & $1.376$       & $2.288$       & $3.053$       & $3.730$       & $2.612$       & $2.244$       & $3.662$ \\ 
        & \textit{Importance}   & $0.35$        & $0.14$        & $0.19$        & $0.11$        & $-0.00$       & $0.25$        & $-0.27$ \\ \midrule 

Data    & \textit{F1 Macro (\%)}        & $90.53$       & $65.30$       & $59.89$       & $43.92$       & $42.76$       & $60.64$       & $60.59$ \\ 
Order   & \textit{F1 Std}       & $3.02$        & $4.23$        & $3.96$        & $6.22$        & $3.67$        & $4.27$        & $4.18$ \\ 
        & \textit{Contributed Std}      & $1.177$       & $2.919$       & $2.910$       & $4.471$       & $2.840$       & $3.600$       & $3.720$ \\ 
        & \textit{Mitigated Std}        & $2.694$       & $2.845$       & $2.383$       & $4.247$       & $2.242$       & $2.123$       & $1.833$ \\ 
        & \textit{Importance}   & $-0.53$       & $0.02$        & $0.11$        & $0.04$        & $0.15$        & $0.25$        & $0.41$ \\ \midrule 

Sample  & \textit{F1 Macro (\%)}        & $89.70$       & $65.54$       & $58.42$       & $45.03$       & $39.89$       & $59.32$       & $62.24$ \\ 
Choice  & \textit{F1 Std}       & $4.69$        & $4.31$        & $5.04$        & $5.92$        & $3.96$        & $4.25$        & $4.09$ \\ 
        & \textit{Contributed Std}      & $3.481$       & $3.630$       & $3.661$       & $5.099$       & $2.783$       & $3.391$       & $2.329$ \\ 
        & \textit{Mitigated Std}        & $1.714$       & $1.911$       & $3.293$       & $2.708$       & $2.775$       & $2.453$       & $3.261$ \\ 
        & \textit{Importance}   & $0.61$        & $0.42$        & $0.08$        & $0.43$        & $0.00$        & $0.16$        & $-0.20$ \\
\bottomrule
\end{tabular}
\end{sc}
\end{small}
\end{center}
\caption{Results from investigating the importance for the effects of different randomness factors for the in-context learning using the LLaMA-2 model across all datasets.}
\label{tab:llama2-results-full}
\end{table*}

\begin{table*}[tbh]
\begin{center}
\begin{small}
\begin{sc}
\begin{tabular}{@{}llccccccc@{}}
\toprule
Mistral-7B      &       & SST2  & CoLA  & MRPC  & AG News       & TREC  & SNIPS & DB-Pedia \\ \midrule 
Golden  & \textit{F1 Macro (\%)} & $67.45$       & $61.96$       & $67.42$       & $65.28$       & $51.66$       & $75.96$       & $90.03$ \\ 
Model   & \textit{F1 Std}       & $13.38$       & $12.73$       & $3.22$        & $6.87$        & $6.37$        & $7.91$        & $2.12$ \\ \midrule 

Label   & \textit{F1 Macro (\%)}        & $66.72$       & $62.30$       & $67.40$       & $64.31$       & $52.54$       & $75.37$       & $89.78$ \\ 
Selection       & \textit{F1 Std}       & $13.79$       & $12.48$       & $3.67$        & $6.30$        & $5.80$        & $8.99$        & $1.87$ \\ 
        & \textit{Contributed Std}      & $10.793$      & $7.913$       & $1.880$       & $4.969$       & $4.360$       & $5.412$       & $1.545$ \\ 
        & \textit{Mitigated Std}        & $6.662$       & $8.511$       & $2.986$       & $3.157$       & $3.626$       & $7.053$       & $0.877$ \\ 
        & \textit{Importance}   & $0.31$        & $-0.05$       & $-0.34$       & $0.26$        & $0.12$        & $-0.21$       & $0.32$ \\ \midrule 

Data    & \textit{F1 Macro (\%)}        & $67.96$       & $64.87$       & $65.40$       & $65.92$       & $51.25$       & $74.19$       & $89.55$ \\ 
Split   & \textit{F1 Std}       & $13.46$       & $12.15$       & $3.59$        & $7.42$        & $6.21$        & $8.17$        & $1.50$ \\ 
        & \textit{Contributed Std}      & $10.935$      & $10.947$      & $2.057$       & $6.018$       & $4.472$       & $6.391$       & $1.161$ \\ 
        & \textit{Mitigated Std}        & $6.302$       & $4.280$       & $2.767$       & $3.871$       & $3.847$       & $4.597$       & $0.677$ \\ 
        & \textit{Importance}   & $0.35$        & $0.52$        & $-0.22$       & $0.31$        & $0.10$        & $0.23$        & $0.23$ \\ \midrule 

Data    & \textit{F1 Macro (\%)}        & $70.31$       & $61.50$       & $66.91$       & $62.94$       & $52.18$       & $77.82$       & $91.09$ \\ 
Order   & \textit{F1 Std}       & $14.97$       & $12.56$       & $3.60$        & $5.58$        & $7.62$        & $7.03$        & $2.67$ \\ 
        & \textit{Contributed Std}      & $8.629$       & $3.018$       & $2.294$       & $3.943$       & $5.626$       & $5.610$       & $1.877$ \\ 
        & \textit{Mitigated Std}        & $10.732$      & $11.459$      & $2.586$       & $3.496$       & $4.846$       & $4.119$       & $1.486$ \\ 
        & \textit{Importance}   & $-0.16$       & $-0.66$       & $-0.09$       & $0.07$        & $0.12$        & $0.19$        & $0.18$ \\ \midrule 

Sample  & \textit{F1 Macro (\%)}        & $66.56$       & $66.78$       & $67.58$       & $64.16$       & $52.58$       & $74.20$       & $90.07$ \\ 
Choice  & \textit{F1 Std}       & $12.78$       & $11.64$       & $3.45$        & $6.96$        & $5.96$        & $7.67$        & $2.32$ \\ 
        & \textit{Contributed Std}      & $12.084$      & $8.865$       & $2.521$       & $6.066$       & $4.306$       & $5.722$       & $1.892$ \\ 
        & \textit{Mitigated Std}        & $3.553$       & $6.853$       & $2.140$       & $3.322$       & $4.051$       & $4.956$       & $0.697$ \\ 
        & \textit{Importance}   & $0.64$        & $0.16$        & $0.12$        & $0.40$        & $0.04$        & $0.10$        & $0.56$ \\
\bottomrule
\end{tabular}
\end{sc}
\end{small}
\end{center}
\caption{Results from investigating the importance for the effects of different randomness factors for the in-context learning using the Mistral-7B model across all datasets.}
\label{tab:mistral-results-full}
\end{table*}

\begin{table*}[tbh]
\begin{center}
\begin{small}
\begin{sc}
\begin{tabular}{@{}llccccccc@{}}
\toprule
Zephyr-7B       &       & SST2  & CoLA  & MRPC  & AG News       & TREC  & SNIPS & DB-Pedia \\ \midrule 
Golden  & \textit{F1 Macro (\%)} & $60.22$       & $51.16$       & $54.74$       & $61.73$       & $59.08$       & $71.73$       & $90.19$ \\ 
Model   & \textit{F1 Std}       & $1.04$        & $9.57$        & $12.79$       & $2.07$        & $3.88$        & $4.13$        & $0.83$ \\ \midrule 

Label   & \textit{F1 Macro (\%)}        & $60.23$       & $48.55$       & $55.29$       & $62.17$       & $58.18$       & $71.65$       & $90.13$ \\ 
Selection       & \textit{F1 Std}       & $1.04$        & $7.27$        & $12.43$       & $1.88$        & $3.52$        & $3.30$        & $0.84$ \\ 
        & \textit{Contributed Std}      & $0.863$       & $5.806$       & $5.109$       & $1.460$       & $2.963$       & $2.935$       & $0.761$ \\ 
        & \textit{Mitigated Std}        & $0.548$       & $2.529$       & $11.008$      & $1.004$       & $1.494$       & $0.977$       & $0.298$ \\ 
        & \textit{Importance}   & $0.30$        & $0.34$        & $-0.46$       & $0.22$        & $0.38$        & $0.47$        & $0.56$ \\ \midrule 

Data    & \textit{F1 Macro (\%)}        & $60.42$       & $50.43$       & $51.84$       & $62.24$       & $57.89$       & $71.76$       & $89.85$ \\ 
Split   & \textit{F1 Std}       & $0.79$        & $9.94$        & $12.42$       & $2.06$        & $3.71$        & $3.99$        & $0.98$ \\ 
        & \textit{Contributed Std}      & $0.664$       & $7.675$       & $5.973$       & $1.817$       & $3.221$       & $3.052$       & $0.823$ \\ 
        & \textit{Mitigated Std}        & $0.380$       & $4.619$       & $10.563$      & $0.807$       & $1.345$       & $2.242$       & $0.466$ \\ 
        & \textit{Importance}   & $0.27$        & $0.32$        & $-0.36$       & $0.49$        & $0.48$        & $0.20$        & $0.43$ \\ \midrule 

Data    & \textit{F1 Macro (\%)}        & $59.97$       & $49.34$       & $55.83$       & $62.56$       & $59.93$       & $71.99$       & $90.12$ \\ 
Order   & \textit{F1 Std}       & $1.05$        & $7.87$        & $10.69$       & $2.01$        & $4.06$        & $3.61$        & $0.85$ \\ 
        & \textit{Contributed Std}      & $0.456$       & $3.598$       & $8.038$       & $0.919$       & $1.925$       & $3.007$       & $0.592$ \\ 
        & \textit{Mitigated Std}        & $0.918$       & $5.379$       & $6.069$       & $1.744$       & $3.550$       & $1.791$       & $0.584$ \\ 
        & \textit{Importance}   & $-0.44$       & $-0.19$       & $0.15$        & $-0.40$       & $-0.42$       & $0.29$        & $0.01$ \\ \midrule 

Sample  & \textit{F1 Macro (\%)}        & $60.13$       & $51.57$       & $52.43$       & $61.97$       & $59.02$       & $70.75$       & $90.26$ \\ 
Choice  & \textit{F1 Std}       & $0.83$        & $9.97$        & $13.69$       & $2.30$        & $3.83$        & $4.08$        & $0.74$ \\ 
        & \textit{Contributed Std}      & $0.744$       & $9.135$       & $6.305$       & $1.874$       & $3.279$       & $3.331$       & $0.576$ \\ 
        & \textit{Mitigated Std}        & $0.338$       & $3.333$       & $11.849$      & $1.144$       & $1.769$       & $2.164$       & $0.433$ \\ 
        & \textit{Importance}   & $0.39$        & $0.61$        & $-0.43$       & $0.35$        & $0.39$        & $0.28$        & $0.17$ \\
\bottomrule
\end{tabular}
\end{sc}
\end{small}
\end{center}
\caption{Results from investigating the importance for the effects of different randomness factors for the in-context learning using the Zephyr-7B model across all datasets.}
\label{tab:zephyr-results-full}
\end{table*}

\begin{table*}[tbh]
\begin{center}
\begin{small}
\begin{sc}
\begin{tabular}{@{}llccccccc@{}}
\toprule
BERT    &       & SST2  & CoLA  & MRPC  & AG News       & TREC  & SNIPS & DB-Pedia \\ \midrule 
Golden  & \textit{F1 Macro (\%)} & $87.37$       & $72.63$       & $73.56$       & $85.78$       & $90.11$       & $97.80$       & $98.80$ \\ 
Model   & \textit{F1 Std}       & $0.97$        & $1.47$        & $2.92$        & $1.24$        & $1.67$        & $0.49$        & $0.36$ \\ \midrule 

Label   & \textit{F1 Macro (\%)}        & $87.29$       & $72.61$       & $73.42$       & $85.79$       & $89.97$       & $97.83$       & $98.81$ \\ 
Selection       & \textit{F1 Std}       & $1.14$        & $1.55$        & $2.76$        & $1.29$        & $1.77$        & $0.51$        & $0.34$ \\ 
        & \textit{Contributed Std}      & $0.927$       & $1.212$       & $2.168$       & $0.979$       & $1.348$       & $0.426$       & $0.308$ \\ 
        & \textit{Mitigated Std}        & $0.453$       & $0.865$       & $1.517$       & $0.776$       & $1.042$       & $0.248$       & $0.121$ \\ 
        & \textit{Importance}   & $0.49$        & $0.24$        & $0.22$        & $0.16$        & $0.18$        & $0.37$        & $0.52$ \\ \midrule 

Data    & \textit{F1 Macro (\%)}        & $87.31$       & $72.43$       & $73.38$       & $85.73$       & $89.54$       & $97.82$       & $98.80$ \\ 
Split   & \textit{F1 Std}       & $1.10$        & $1.40$        & $2.90$        & $1.27$        & $1.71$        & $0.48$        & $0.32$ \\ 
        & \textit{Contributed Std}      & $0.937$       & $1.250$       & $2.384$       & $1.164$       & $1.568$       & $0.442$       & $0.294$ \\ 
        & \textit{Mitigated Std}        & $0.361$       & $0.528$       & $1.436$       & $0.388$       & $0.523$       & $0.142$       & $0.115$ \\ 
        & \textit{Importance}   & $0.59$        & $0.49$        & $0.33$        & $0.63$        & $0.63$        & $0.62$        & $0.50$ \\ \midrule 

Model   & \textit{F1 Macro (\%)}        & $87.31$       & $72.59$       & $73.50$       & $85.79$       & $90.30$       & $97.64$       & $98.84$ \\ 
Initialisation  & \textit{F1 Std}       & $1.12$        & $1.52$        & $2.81$        & $1.18$        & $1.79$        & $0.49$        & $0.33$ \\ 
        & \textit{Contributed Std}      & $0.828$       & $1.059$       & $2.180$       & $0.693$       & $0.939$       & $0.270$       & $0.121$ \\ 
        & \textit{Mitigated Std}        & $0.512$       & $1.000$       & $1.600$       & $0.903$       & $1.491$       & $0.387$       & $0.300$ \\ 
        & \textit{Importance}   & $0.33$        & $0.04$        & $0.20$        & $-0.17$       & $-0.33$       & $-0.24$       & $-0.50$ \\ \midrule 

Data    & \textit{F1 Macro (\%)}        & $87.30$       & $72.64$       & $73.66$       & $85.79$       & $90.26$       & $97.64$       & $98.84$ \\ 
Order   & \textit{F1 Std}       & $1.03$        & $1.63$        & $2.80$        & $1.10$        & $1.76$        & $0.47$        & $0.32$ \\ 
        & \textit{Contributed Std}      & $0.852$       & $1.086$       & $2.081$       & $0.686$       & $1.019$       & $0.246$       & $0.103$ \\ 
        & \textit{Mitigated Std}        & $0.417$       & $1.151$       & $1.604$       & $0.817$       & $1.371$       & $0.392$       & $0.301$ \\ 
        & \textit{Importance}   & $0.45$        & $-0.04$       & $0.16$        & $-0.11$       & $-0.21$       & $-0.30$       & $-0.55$ \\
\bottomrule
\end{tabular}
\end{sc}
\end{small}
\end{center}
\caption{Results from investigating the importance for the effects of different randomness factors for the BERT fine-tuning across all datasets.}
\label{tab:bert-results-full}
\end{table*}

\begin{table*}[tbh]
\begin{center}
\begin{small}
\begin{sc}
\begin{tabular}{@{}llccccccc@{}}
\toprule
RoBERTa &       & SST2  & CoLA  & MRPC  & AG News       & TREC  & SNIPS & DB-Pedia \\ \midrule 
Golden  & \textit{F1 Macro (\%)} & $88.48$       & $74.60$       & $80.35$       & $86.49$       & $91.66$       & $98.16$       & $98.31$ \\ 
Model   & \textit{F1 Std}       & $1.29$        & $3.22$        & $2.16$        & $1.56$        & $1.79$        & $0.58$        & $0.57$ \\ \midrule 

Label   & \textit{F1 Macro (\%)}        & $88.54$       & $74.57$       & $80.25$       & $86.66$       & $91.55$       & $98.16$       & $98.35$ \\ 
Selection       & \textit{F1 Std}       & $1.05$        & $3.54$        & $2.10$        & $1.35$        & $1.72$        & $0.56$        & $0.69$ \\ 
        & \textit{Contributed Std}      & $0.904$       & $2.171$       & $1.723$       & $1.150$       & $1.461$       & $0.455$       & $0.471$ \\ 
        & \textit{Mitigated Std}        & $0.392$       & $1.312$       & $0.990$       & $0.639$       & $0.732$       & $0.243$       & $0.234$ \\ 
        & \textit{Importance}   & $0.40$        & $0.27$        & $0.34$        & $0.33$        & $0.41$        & $0.36$        & $0.41$ \\ \midrule 

Data    & \textit{F1 Macro (\%)}        & $88.45$       & $74.24$       & $80.13$       & $86.54$       & $91.15$       & $98.10$       & $98.37$ \\ 
Split   & \textit{F1 Std}       & $1.21$        & $3.51$        & $2.20$        & $1.48$        & $1.75$        & $0.57$        & $0.44$ \\ 
        & \textit{Contributed Std}      & $0.992$       & $2.151$       & $1.981$       & $1.377$       & $1.581$       & $0.506$       & $0.398$ \\ 
        & \textit{Mitigated Std}        & $0.375$       & $1.162$       & $0.709$       & $0.392$       & $0.596$       & $0.168$       & $0.164$ \\ 
        & \textit{Importance}   & $0.48$        & $0.31$        & $0.59$        & $0.63$        & $0.55$        & $0.58$        & $0.41$ \\ \midrule 

Model   & \textit{F1 Macro (\%)}        & $88.53$       & $74.57$       & $80.29$       & $86.59$       & $91.48$       & $98.02$       & $98.40$ \\ 
Initialisation  & \textit{F1 Std}       & $1.10$        & $3.95$        & $2.16$        & $1.49$        & $1.80$        & $0.60$        & $0.42$ \\ 
        & \textit{Contributed Std}      & $0.890$       & $2.051$       & $1.552$       & $1.030$       & $1.380$       & $0.412$       & $0.234$ \\ 
        & \textit{Mitigated Std}        & $0.457$       & $1.705$       & $1.312$       & $0.953$       & $1.038$       & $0.367$       & $0.321$ \\ 
        & \textit{Importance}   & $0.34$        & $0.11$        & $0.11$        & $0.05$        & $0.19$        & $0.08$        & $-0.15$ \\ \midrule 

Data    & \textit{F1 Macro (\%)}        & $88.42$       & $74.35$       & $80.40$       & $86.71$       & $91.52$       & $98.06$       & $98.38$ \\ 
Order   & \textit{F1 Std}       & $1.26$        & $4.35$        & $2.10$        & $1.24$        & $1.81$        & $0.58$        & $0.41$ \\ 
        & \textit{Contributed Std}      & $1.033$       & $2.424$       & $1.649$       & $0.991$       & $1.312$       & $0.372$       & $0.223$ \\ 
        & \textit{Mitigated Std}        & $0.447$       & $1.769$       & $1.097$       & $0.671$       & $1.168$       & $0.412$       & $0.333$ \\ 
        & \textit{Importance}   & $0.46$        & $0.20$        & $0.26$        & $0.20$        & $0.08$        & $-0.07$       & $-0.19$ \\
\bottomrule
\end{tabular}
\end{sc}
\end{small}
\end{center}
\caption{Results from investigating the importance for the effects of different randomness factors for the RoBERTa fine-tuning across all datasets.}
\label{tab:roberta-results-full}
\end{table*}

\begin{table*}[tbh]
\begin{center}
\begin{small}
\begin{sc}
\begin{tabular}{@{}llccc@{}}
\toprule
Prototypical Networks   &       & SST2  & CoLA  & MRPC \\ \midrule 
Golden  & \textit{F1 Macro(\%)} & $80.33$       & $60.70$       & $63.62$ \\ 
Model   & \textit{F1 Std}       & $0.94$        & $2.11$        & $1.78$ \\ \midrule 

Label   & \textit{F1 Macro (\%)}        & $80.33$       & $60.65$       & $63.34$ \\ 
Selection       & \textit{F1 Std}       & $1.04$        & $2.10$        & $1.57$ \\ 
        & \textit{Contributed Std}      & $0.959$       & $1.924$       & $1.448$ \\ 
        & \textit{Mitigated Std}        & $0.268$       & $0.665$       & $0.472$ \\ 
        & \textit{Importance}   & $0.74$        & $0.60$        & $0.55$ \\ \midrule 

Data    & \textit{F1 Macro (\%)}        & $80.35$       & $60.23$       & $63.21$ \\ 
Split   & \textit{F1 Std}       & $0.97$        & $2.18$        & $1.91$ \\ 
        & \textit{Contributed Std}      & $0.887$       & $2.010$       & $1.791$ \\ 
        & \textit{Mitigated Std}        & $0.283$       & $0.646$       & $0.508$ \\ 
        & \textit{Importance}   & $0.64$        & $0.65$        & $0.72$ \\ \midrule 

Model   & \textit{F1 Macro (\%)}        & $80.20$       & $61.04$       & $63.09$ \\ 
Initialisation  & \textit{F1 Std}       & $0.97$        & $2.22$        & $1.88$ \\ 
        & \textit{Contributed Std}      & $0.887$       & $1.801$       & $1.240$ \\ 
        & \textit{Mitigated Std}        & $0.631$       & $1.186$       & $1.348$ \\ 
        & \textit{Importance}   & $0.27$        & $0.29$        & $-0.06$ \\ \midrule 

Data    & \textit{F1 Macro (\%)}        & $75.77$       & $59.80$       & $62.98$ \\ 
Order   & \textit{F1 Std}       & $4.51$        & $4.11$        & $3.08$ \\ 
        & \textit{Contributed Std}      & $3.233$       & $3.346$       & $2.265$ \\ 
        & \textit{Mitigated Std}        & $2.371$       & $1.659$       & $1.412$ \\ 
        & \textit{Importance}   & $0.92$        & $0.80$        & $0.48$ \\ \midrule 

Sample  & \textit{F1 Macro (\%)}        & $80.41$       & $60.54$       & $63.30$ \\ 
Choice  & \textit{F1 Std}       & $0.98$        & $2.16$        & $1.80$ \\ 
        & \textit{Contributed Std}      & $0.646$       & $1.659$       & $1.084$ \\ 
        & \textit{Mitigated Std}        & $0.630$       & $1.335$       & $1.393$ \\ 
        & \textit{Importance}   & $0.02$        & $0.15$        & $-0.17$ \\
\bottomrule
\end{tabular}
\end{sc}
\end{small}
\end{center}
\caption{Results from investigating the importance for the effects of different randomness factors for the Prototypical Networks meta-learning approach across all binary datasets.}
\label{tab:protonet-results-full}
\end{table*}

\begin{table*}[tbh]
\begin{center}
\begin{small}
\begin{sc}
\begin{tabular}{@{}llccc@{}}
\toprule
MAML    &       & SST2  & CoLA  & MRPC \\ \midrule 
Golden  & \textit{F1 Macro(\%)} & $79.93$       & $60.18$       & $58.29$ \\ 
Model   & \textit{F1 Std}       & $2.34$        & $1.86$        & $6.27$ \\ \midrule 

Label   & \textit{F1 Macro (\%)}        & $79.99$       & $60.02$       & $57.52$ \\ 
Selection       & \textit{F1 Std}       & $1.27$        & $1.84$        & $6.55$ \\ 
        & \textit{Contributed Std}      & $0.893$       & $1.706$       & $6.000$ \\ 
        & \textit{Mitigated Std}        & $0.500$       & $0.512$       & $1.988$ \\ 
        & \textit{Importance}   & $0.17$        & $0.64$        & $0.64$ \\ \midrule 

Data    & \textit{F1 Macro (\%)}        & $80.19$       & $59.95$       & $57.72$ \\ 
Split   & \textit{F1 Std}       & $0.95$        & $1.86$        & $6.60$ \\ 
        & \textit{Contributed Std}      & $0.819$       & $1.716$       & $5.868$ \\ 
        & \textit{Mitigated Std}        & $0.286$       & $0.555$       & $2.188$ \\ 
        & \textit{Importance}   & $0.23$        & $0.62$        & $0.59$ \\ \midrule 

Model   & \textit{F1 Macro (\%)}        & $79.98$       & $60.76$       & $57.98$ \\ 
Initialisation  & \textit{F1 Std}       & $1.67$        & $1.98$        & $5.67$ \\ 
        & \textit{Contributed Std}      & $0.678$       & $1.389$       & $4.792$ \\ 
        & \textit{Mitigated Std}        & $0.897$       & $1.328$       & $2.288$ \\ 
        & \textit{Importance}   & $-0.09$       & $0.03$        & $0.40$ \\ \midrule 

Data    & \textit{F1 Macro (\%)}        & $79.58$       & $59.17$       & $55.00$ \\ 
Order   & \textit{F1 Std}       & $1.54$        & $2.96$        & $10.85$ \\ 
        & \textit{Contributed Std}      & $1.010$       & $2.368$       & $9.522$ \\ 
        & \textit{Mitigated Std}        & $0.827$       & $1.340$       & $3.727$ \\ 
        & \textit{Importance}   & $0.08$        & $0.55$        & $0.92$ \\ \midrule 

Sample  & \textit{F1 Macro (\%)}        & $80.19$       & $60.04$       & $58.10$ \\ 
Choice  & \textit{F1 Std}       & $1.00$        & $1.89$        & $6.36$ \\ 
        & \textit{Contributed Std}      & $0.167$       & $1.265$       & $1.940$ \\ 
        & \textit{Mitigated Std}        & $0.977$       & $1.352$       & $5.983$ \\ 
        & \textit{Importance}   & $-0.35$       & $-0.05$       & $-0.64$ \\
\bottomrule
\end{tabular}
\end{sc}
\end{small}
\end{center}
\caption{Results from investigating the importance for the effects of different randomness factors for the MAML meta-learning approach across all binary datasets.}
\label{tab:maml-results-full}
\end{table*}

\begin{table*}[tbh]
\begin{center}
\begin{small}
\begin{sc}
\begin{tabular}{@{}llccc@{}}
\toprule
Reptile &       & SST2  & CoLA  & MRPC \\ \midrule 
Golden  & \textit{F1 Macro(\%)} & $81.14$       & $57.17$       & $61.06$ \\ 
Model   & \textit{F1 Std}       & $1.46$        & $10.50$       & $5.70$ \\ \midrule 

Label   & \textit{F1 Macro (\%)}        & $81.06$       & $56.16$       & $60.30$ \\ 
Selection       & \textit{F1 Std}       & $0.93$        & $11.08$       & $5.89$ \\ 
        & \textit{Contributed Std}      & $0.897$       & $9.482$       & $4.745$ \\ 
        & \textit{Mitigated Std}        & $0.141$       & $3.398$       & $1.819$ \\ 
        & \textit{Importance}   & $0.52$        & $0.58$        & $0.51$ \\ \midrule 

Data    & \textit{F1 Macro (\%)}        & $81.04$       & $56.45$       & $60.54$ \\ 
Split   & \textit{F1 Std}       & $1.56$        & $10.24$       & $5.92$ \\ 
        & \textit{Contributed Std}      & $0.747$       & $8.550$       & $4.740$ \\ 
        & \textit{Mitigated Std}        & $0.485$       & $3.175$       & $1.776$ \\ 
        & \textit{Importance}   & $0.18$        & $0.51$        & $0.52$ \\ \midrule 

Model   & \textit{F1 Macro (\%)}        & $81.01$       & $56.87$       & $59.84$ \\ 
Initialisation  & \textit{F1 Std}       & $2.42$        & $10.65$       & $6.73$ \\ 
        & \textit{Contributed Std}      & $0.576$       & $8.325$       & $4.853$ \\ 
        & \textit{Mitigated Std}        & $1.300$       & $3.979$       & $2.722$ \\ 
        & \textit{Importance}   & $-0.50$       & $0.41$        & $0.37$ \\ \midrule 

Data    & \textit{F1 Macro (\%)}        & $81.17$       & $59.17$       & $39.87$ \\ 
Order   & \textit{F1 Std}       & $2.01$        & $7.34$        & $12.33$ \\ 
        & \textit{Contributed Std}      & $0.591$       & $4.690$       & $11.446$ \\ 
        & \textit{Mitigated Std}        & $0.951$       & $2.959$       & $3.991$ \\ 
        & \textit{Importance}   & $-0.25$       & $0.16$        & $1.31$ \\ \midrule 

Sample  & \textit{F1 Macro (\%)}        & $81.00$       & $60.00$       & $60.77$ \\ 
Choice  & \textit{F1 Std}       & $2.61$        & $4.97$        & $5.63$ \\ 
        & \textit{Contributed Std}      & $0.370$       & $2.510$       & $4.221$ \\ 
        & \textit{Mitigated Std}        & $1.674$       & $2.171$       & $2.612$ \\ 
        & \textit{Importance}   & $-0.89$       & $0.03$        & $0.28$ \\
\bottomrule
\end{tabular}
\end{sc}
\end{small}
\end{center}
\caption{Results from investigating the importance for the effects of different randomness factors for the Reptile meta-learning approach across all binary datasets.}
\label{tab:reptile-results-full}
\end{table*}

\end{document}